\documentclass{article}
\pdfoutput=1
\usepackage[round]{natbib}
\usepackage{fullpage}
\usepackage{times}

\usepackage[utf8]{inputenc}
\usepackage[T1]{fontenc}
\usepackage[hyphens]{url}
\usepackage{graphicx}
\usepackage{caption}
\usepackage{subcaption}

\usepackage{algorithm}
\usepackage{algorithmic}
\usepackage{booktabs}
\usepackage{amsmath}
\usepackage{amssymb}
\usepackage{amsfonts}
\usepackage{mathtools}
\usepackage{amsthm}
\usepackage{nicefrac}
\usepackage{microtype}
\usepackage{xcolor}
\usepackage{wrapfig}
\definecolor{revorange}{RGB}{230,110,0}
\definecolor{revblue}{RGB}{20,70,200}

\usepackage{hyperref}
\usepackage[capitalize,noabbrev]{cleveref}
\usepackage[textsize=tiny]{todonotes}
\usepackage{tikz}
\usetikzlibrary{arrows.meta,shapes.geometric,calc,backgrounds,positioning}

\theoremstyle{plain}
\newtheorem{theorem}{Theorem}
\newtheorem{proposition}{Proposition}
\newtheorem{lemma}{Lemma}
\newtheorem{corollary}{Corollary}
\theoremstyle{definition}

\newtheorem{assumption}{Assumption}
\theoremstyle{remark}
\newtheorem{remark}{Remark}

\title{Global Optimization on Graph-Structured Data via Gaussian Processes with Spectral Representations}

\author{%
Shu Hong\thanks{George Washington University, United States. E-mail: \texttt{shu.hong@gwu.edu}, \texttt{tlan@gwu.edu}.}
\quad
Yongsheng Mei\thanks{Amazon. E-mail: \texttt{ysmei97@gmail.com}.}
\quad
Mahdi Imani\thanks{Northeastern University, United States. E-mail: \texttt{m.imani@northeastern.edu}.}
\quad
Tian Lan\footnotemark[1]
}

\date{}
\begin{document}
\maketitle
\begin{abstract}
Bayesian optimization (BO) is a powerful framework for optimizing expensive black-box objectives, yet extending it to graph-structured domains remains challenging due to the discrete and combinatorial nature of graphs. Existing approaches often rely either on full graph topology, which is impractical for large or partially observed graphs, or on incremental local exploration, which can lead to slow convergence. We introduce a scalable framework for global optimization over graphs that builds Gaussian process (GP) surrogates from low-rank spectral representations inferred from sparse structural observations. The method fits a low-rank graph surrogate to the observed connections and uses its spectrum to define a kernel over all nodes, so that function evaluations and structural observations jointly refine a global model, enabling efficient global search and principled uncertainty estimation even with limited data. We further show how a limited budget of structural observations can be allocated according to its expected value for the optimization. We also provide theoretical analysis establishing conditions under which the surrogate geometry is identified under different sampling regimes, and bounding the optimization error incurred under an estimated geometry. Experiments on synthetic and real-world datasets demonstrate that our approach achieves lower regret and improved optimization performance compared to prior methods.
\end{abstract}

\section{Introduction}
Many real-world decision-making problems involve optimizing an unknown function defined over the nodes of a large graph. In such settings, each node represents a discrete action, configuration, or intervention, while querying the function value at a node is often costly, either via simulation, experimentation, or human evaluation. The goal is therefore to identify the node with the optimal function value using as few function evaluations as possible. This problem arises in diverse applications, including selecting molecules in a similarity graph~\citep{korovina2020chembo}, prioritizing genes
in a protein interaction network~\citep{tsherniak2017defining,huttlin2021dual}, and targeting accounts
in a social network~\citep{mcauley2012socialcircles}. Across these domains, exhaustive evaluation over all nodes is infeasible, which necessitates sample-efficient optimization that exploits structural correlations induced by the graph topology.

Bayesian optimization (BO)~\citep{frazier2018tutorial,snoek2012practical} provides a principled framework for such sample-efficient optimization. It employs a probabilistic surrogate, typically a Gaussian process (GP), to guide exploration via an acquisition function. A fundamental requirement for effective BO on graphs is a globally defined kernel that propagates uncertainty and structural correlations across the entire search space. Existing methods typically construct these kernels from the graph Laplacian or spectral embeddings, implicitly assuming that the graph topology is fully known a priori~\citep{cui2019deep,ru2020interpretable,oh2019combinatorial,xie2025global}.

However, in many practical settings, this assumption of full observability does not hold. Graph structures are often incomplete, costly to acquire, or revealed only incrementally through limited edge queries. We therefore study BO on graphs with partially observed topology. At each iteration, the learner performs one expensive node evaluation and, in parallel, queries a limited budget of adjacency entries or entire rows to refine its knowledge of the graph. Recent graph-specific BO methods~\citep{wan2023bayesian} address this by incremental local exploration of the neighborhood. Without a global geometric representation, posterior uncertainty cannot propagate beyond the revealed subgraph, so exploration is myopic and no existing method maintains a global posterior and acquisition rule over all nodes.

We propose a novel framework for global graph BO that enables sample-efficient optimization on large and partially observed graphs. Our core insight is that effective global BO under partial topology requires inferring a coherent surrogate geometry rather than discovering structure neighborhood by neighborhood. To this end, we fit a graph-valid low-rank surrogate to the revealed entries, extract its Laplacian spectral structure efficiently to obtain a global geometry, and use this geometry to define a GP surrogate and acquisition function over all nodes. The same geometry determines what each budget can learn. On the surrogate the GP posterior is a linear model in $d$-dimensional node features with a coefficient shared across nodes, so function evaluations learn the coefficient and edge queries learn the features. Structure acquisition therefore becomes a decision problem: we value an edge query by its expected effect on the next selected node, which yields a knowledge-gradient rule for pair queries and a simple pin-then-uniform rule for row queries.

We provide theoretical guarantees that characterize when a globally coherent surrogate graph geometry can be inferred from partial edge observations and what an estimated geometry costs the optimizer. For graphs in the surrogate class, $d+1$ revealed rows identify the surrogate while $d$ do not suffice in general. A per-round bound charges the estimated geometry only at the selected node and the optimum, and the resulting simple-regret bound depends on structural access only through the identification time. Experiments on protein interaction networks, a molecular similarity graph, a Wikipedia page graph, and synthetic graphs answer three questions at matched node and edge budgets: the global surrogate has lower final regret than random search, local search, and local graph BO on all ten objectives, full-information controls separate the representation from its estimation, and the knowledge-gradient rule has lower final regret than uniform pairs at every budget.

Our main contributions are summarized as follows:
\begin{itemize}
  \item \textbf{Global BO under partial topology.}
  We formulate Bayesian optimization on graph nodes when topology is incomplete and revealed through budgeted edge queries of adjacency entries or rows, supporting globally defined uncertainty propagation and acquisition over all nodes.

  \item \textbf{Surrogate geometry via a graph-valid low-rank class and its exact spectrum.} We construct a globally defined GP kernel by fitting a graph-valid low-rank surrogate to the revealed entries and exploiting its exactly computed Laplacian spectral geometry, which makes the GP a linear model in node features.

  \item \textbf{Decision-aware structure acquisition and identification guarantees.}
  We price each edge query by its expected effect on the next decision, yielding a knowledge-gradient rule for pair queries and a pin-then-uniform rule for rows, and we prove that $d+1$ rows identify the surrogate while $d$ do not suffice in general, with a simple-regret bound in which structural access enters only through the identification time.

  \item \textbf{Empirical validation on real and synthetic graphs.}
  We demonstrate lower final regret than random search, local search, and local graph BO on protein interaction networks, a molecular similarity graph, a Wikipedia page graph, and synthetic graphs under the same node and edge query budgets.
\end{itemize}

\section{Related Work}

\paragraph{Bayesian Optimization on Graphs.}
BO on graphs places a GP prior on the nodes through a graph kernel or an embedding~\citep{wan2023bayesian,cui2018graph,cui2019deep,ru2020interpretable}, or exploits an explicit combinatorial construction such as a product graph~\citep{oh2019combinatorial,ramachandram2018bayesian,white2021bananas,xie2025global,liang2024bayesian,yuan2024graph,xie2025bogrape}, and assumes the topology known. The closest analysis is the spectral bandit~\citep{valko2014spectral}, a linear bandit in the Laplacian eigenbasis of a known graph with regret governed by an effective dimension. Our GP is such a linear model, but its basis is learned from budgeted edge queries that the learner chooses. Under partial topology the closest prior method explores the revealed neighborhood locally, with row access at every visited node~\citep{wan2023bayesian}.

\paragraph{Low-Rank Graph Models and Structure Recovery.}
Matrix completion recovers a low-rank matrix from uniform samples under incoherence~\citep{candes2009exact,candes2010power,keshavan2010matrix} or from deterministic patterns~\citep{lee2023matrixcompletiongeneraldeterministic}, and adaptive row or column sampling removes the incoherence requirement and the logarithmic factor~\citep{krishnamurthy2013lowrank,mahoney2009cur}. Our surrogate class is the step-graphon or mixed-membership model~\citep{lovasz2012large,hoff2002latent,airoldi2008mixed,karrer2011stochastic}, whose spectral estimation is well understood on a fully observed graph~\citep{chatterjee2015usvt,xu2018rates,mao2021estimating}. Recovery of $A$ is not our goal. The class is what makes every spectral kernel of the surrogate a floor plus a rank-$d$ term, and its landmark structure is what makes a revealed entry a linear measurement of one node, enabling decision-aware allocation of the edge budget.

\paragraph{Spectral Kernels, Uncertain Inputs, and Value of Information.}
Laplacian and diffusion kernels give GP priors on fully observed graphs~\citep{imre2002diffusion,smola2003kernels,borovitskiy2021matern}, and regression on Laplacian eigenmaps is their finite-rank form~\citep{belkin2003laplacian,belkin2004semi}. GPs with uncertain inputs~\citep{girard2003gaussian,mchutchon2011gaussian} and BO under input uncertainty~\citep{oliveira2019bayesian} take the input distribution as given, whereas here the input is a feature whose uncertainty the learner can reduce by a query. Selecting structural queries is a value-of-information problem, connecting our method to the knowledge gradient~\citep{frazier2008knowledge,frazier2010paradoxes} and multi-fidelity BO~\citep{kandasamy2016multi}. Here, budgeted structural observations refine the kernel geometry used for objective prediction.

\section{Problem Statement}
\label{sec:problem}
We consider the problem of optimizing an expensive-to-evaluate black-box function defined over nodes of a large, partially observed graph. Formally, let $G=(\mathcal{V},\mathcal{E})$ denote an undirected and weighted graph, where $\mathcal{V}=\{1,\dots,n\}$ is the node set, and edges $\mathcal{E}=\{(i,j)\in \mathcal{V}\times \mathcal{V} \mid A_{ij}>0\}$ are determined by an unknown symmetric adjacency matrix $A \in [0,1]^{n\times n}$.
$A_{ij}>0$ indicates that nodes $i$ and $j$ are connected and $A_{ij}=0$ otherwise.
A black-box function $y:\mathcal{V}\rightarrow \mathbb{R}$ is defined on the graph nodes.

Evaluating $y(v)$ is expensive, so we can query $y(v)$ only at a small subset of nodes.
Our goal is to identify the globally optimal node
\begin{equation}
\label{Opt:black-box problem}
v^*=\arg\max_{v\in \mathcal{V}} y(v),
\end{equation}
using as few function evaluations as possible.

BO proceeds iteratively. Let $\mathcal V^{(t)}\subseteq\mathcal V$ be the set of nodes evaluated up to iteration $t$. From the observed pairs $\{(v,y(v))\}_{v\in\mathcal V^{(t)}}$ it forms a posterior over $y$ and evaluates the node $v^{(t+1)}$ that maximizes an acquisition function, which trades off exploration and exploitation under that posterior. We consider exact observations of $y$ and measure performance by the simple regret $r_T=y(v^*)-\max_{v\in\mathcal V^{(T)}}y(v)$ after $T$ iterations.

In parallel with node function evaluations, the learner acquires the graph structure through \emph{edge queries}, which are limited by a per-iteration budget. Depending on the application, we consider two query types:
(i) A \emph{pair query} reveals a single entry $A_{uv}$ of a chosen node pair, and the learner may issue $Q_E$ such queries per iteration. This is the natural unit when affinities are computed or assayed one pair at a time, such as similarities between molecules~\citep{rogers2010extended,bajusz2015tanimoto}. 
(ii) A \emph{row query} reveals all entries $A_{u,\cdot}$ of a chosen node $u$, and the learner may issue $Q_R$ such queries per iteration. This is how relational data arrive when one measurement returns all partners of one entity, e.g., an affinity-purification experiment with bait protein $u$~\citep{huttlin2021dual}, or a call to a social-network interface that returns the contact list of one account~\citep{mcauley2012socialcircles}. 
Let $\Omega^{(t)}\subseteq\mathcal V\times\mathcal V$ be the set of node pairs revealed up to iteration $t$, and let $A_\Omega^{(t)}=\mathcal P_{\Omega^{(t)}}(A)$ be the partially observed adjacency, where $\mathcal P_\Omega$ keeps the entries in $\Omega$ and sets the others to zero. Self-pairs are never queried, and we write $A_{\mathrm{off}}$ for $A$ with its diagonal removed, the object the queries can reveal. Revealed entries are exact, and a revealed zero $A_{ij}=0$ is itself an observation, since it rules out an edge.

\begin{figure}[t]
\centering
\resizebox{0.9\linewidth}{!}{%
\begin{tikzpicture}[
  font=\footnotesize, x=1cm, y=1cm,
  panel/.style={draw=black!22,rounded corners=3pt,line width=0.4pt},
  title/.style={font=\footnotesize\bfseries,anchor=north west,inner sep=2pt},
  lab/.style={font=\scriptsize,inner sep=1pt},
  tlab/.style={font=\tiny,inner sep=1pt},
  val/.style={circle,draw=blue!22!black,line width=0.22pt,inner sep=0pt,minimum size=5.5pt},
  ev/.style={val,draw=black,line width=1.15pt,minimum size=6.2pt},
  unk/.style={circle,draw=black!35,fill=white,inner sep=0pt,minimum size=5pt,line width=0.35pt},
  halo/.style={draw=purple!70!black,densely dashed,fill=purple!8,line width=0.5pt},
  obs/.style={draw=black!80,line width=0.9pt},
  zero/.style={draw=black!55,line width=0.38pt,densely dotted},
  pair/.style={draw=black!80,line width=0.9pt},
  unobs/.style={draw=black!20,line width=0.35pt,dashed},
  inf/.style={draw=black!30,line width=0.55pt},
  infw/.style={draw=black!16,line width=0.35pt},
  nextnode/.style={star,star points=5,star point ratio=2.2,fill=red!80!black,draw=red!50!black,inner sep=0pt,minimum size=9pt},
  bigarr/.style={-{Stealth[length=2.2mm]},line width=1pt,draw=black!65},
  qarr/.style={-{Stealth[length=2mm]},line width=1.1pt,draw=teal!70!black},
  sbar/.style={line width=3.2pt,draw=black!60},
]
\begin{scope}
\draw[panel] (0,0) rectangle (4.5,4.2);
\node[title] at (0,4.2) {(a) Partial observations};
\coordinate (n1) at (0.45,3.0);  \coordinate (n2) at (1.25,3.3);  \coordinate (n3) at (2.15,3.1);
\coordinate (n4) at (3.25,3.25); \coordinate (n5) at (4.05,2.65); \coordinate (n6) at (0.6,2.1);
\coordinate (n7) at (1.7,2.3);   \coordinate (n8) at (2.7,2.2);   \coordinate (n9) at (3.6,1.8);
\coordinate (n10) at (0.95,1.15);\coordinate (n11) at (1.95,0.9); \coordinate (n12) at (3.0,0.8);
\coordinate (n13) at (3.95,0.65);\coordinate (n14) at (0.4,0.85);
\foreach \t in {n2,n3,n8,n10} \draw[obs] (n7)--(\t);
\foreach \t in {n1,n4,n5,n6,n9,n11,n12,n13,n14} \draw[zero] (n7)--(\t);
\draw[pair] (n4)--(n5) node[tlab,midway,above right=-1pt,text=black!65] {0.6};
\draw[pair] (n11)--(n12) node[tlab,midway,below=-1pt,text=black!65] {0.2};
\draw[zero] (n9)--(n13) node[tlab,midway,right=-1pt,text=black!55] {0};
\draw[pair] (n6)--(n14);
\draw[unobs] (n1)--(n4) node[tlab,midway,above,text=black!35] {?};
\draw[unobs] (n3)--(n9) node[tlab,midway,right,text=black!35] {?};
\foreach \p in {n1,n3,n4,n5,n6,n8,n9,n11,n13,n14} \node[unk] at (\p) {};
\node[ev,fill=blue!80] at (n2) {}; \node[ev,fill=blue!30] at (n10) {}; \node[ev,fill=blue!50] at (n12) {};
\node[unk] at (n7) {};
\node[tlab,anchor=south west,text=blue!50!black] at ($(n2)+(0.06,0.05)$) {$y{=}0.8$};
\node[tlab,anchor=north east,text=blue!50!black] at ($(n10)+(-0.04,-0.05)$) {$y{=}0.3$};
\node[tlab,anchor=north west,text=blue!50!black] at ($(n12)+(0.05,-0.05)$) {$y{=}0.5$};
\node[tlab,anchor=west,text=black!65] at ($(n6)!0.5!(n14)+(0.08,0)$) {pair query};
\node[tlab,anchor=south west] at (0.08,0.25) {\tikz\shade[left color=blue!10,right color=blue!85](0,0)rectangle(0.5,0.13); observed $y$};
\node[tlab,anchor=south west] at (2.45,0.25) {\tikz\draw[obs](0,0)--(0.25,0); observed edge};
\node[tlab,anchor=south west] at (0.08,0.04) {\tikz\draw[zero](0,0)--(0.25,0); observed zero};
\node[tlab,anchor=south west] at (2.45,0.04) {\tikz\draw[unobs](0,0)--(0.25,0); unobserved};
\end{scope}
\begin{scope}[shift={(4.9,0)}]
\draw[panel] (0,0) rectangle (4.9,4.2);
\node[title] at (0,4.2) {(b) Global predictions};
\coordinate (bA) at (3.45,3.05); \coordinate (bB) at (3.95,2.05); \coordinate (bE8) at (3.05,2.55); \coordinate (bH1) at (2.7,3.05); \coordinate (bH2) at (3.9,3.2);
\coordinate (bL) at (1.3,2.95); \coordinate (bE5) at (1.85,2.6); \coordinate (bM1) at (0.75,3.3); \coordinate (bM2) at (1.05,2.35);
\coordinate (bE3) at (1.2,1.25); \coordinate (bL1) at (0.7,0.85); \coordinate (bL2) at (1.7,0.7); \coordinate (bL3) at (2.3,1.7); \coordinate (bL4) at (2.5,1.0);
\foreach \u/\v in {bA/bB,bA/bE8,bA/bH1,bA/bH2,bB/bE8,bB/bH2,bE8/bH1,bH1/bH2,bE8/bH2, bL/bE5,bL/bM1,bL/bM2,bE5/bM2,bM1/bM2, bE3/bL1,bE3/bL2,bE3/bL3,bE3/bL4,bL1/bL2,bL2/bL4,bL3/bL4} \draw[inf] (\u)--(\v);
\foreach \u/\v in {bE5/bH1,bL3/bE8,bM2/bE3,bL/bA} \draw[infw] (\u)--(\v);
\draw[halo] (3.95,2.05) ellipse (0.42 and 0.27);
\draw[halo] (0.75,3.3) ellipse (0.3 and 0.2);
\draw[halo] (0.7,0.85) ellipse (0.34 and 0.22);
\node[ev,fill=blue!80] at (3.05,2.55) {};
\node[val,fill=blue!60] at (2.7,3.05) {};
\node[val,fill=blue!65] at (3.9,3.2) {};
\node[val,fill=blue!78] (A) at (3.45,3.05) {};
\node[lab,anchor=south west,inner sep=0.5pt] at (3.5,3.1) {$\mathrm{A}$};
\node[val,fill=blue!70] (B) at (3.95,2.05) {};
\node[lab,anchor=west,inner sep=0.5pt] at (4.42,2.05) {$\mathrm{B}$};
\node[val,fill=blue!45] at (1.3,2.95) {};
\node[ev,fill=blue!50] at (1.85,2.6) {};
\node[val,fill=blue!42] at (0.75,3.3) {};
\node[val,fill=blue!46] at (1.05,2.35) {};
\node[ev,fill=blue!30] at (1.2,1.25) {};
\node[val,fill=blue!25] at (0.7,0.85) {};
\node[val,fill=blue!22] at (1.7,0.7) {};
\node[val,fill=blue!35] at (2.3,1.7) {};
\node[val,fill=blue!28] at (2.5,1.0) {};
\node[anchor=west,inner sep=0pt] at (0.10,0.42) {\tikz\shade[left color=blue!10,right color=blue!85](0,0)rectangle(0.5,0.13);};
\node[tlab,anchor=west] at (0.74,0.42) {predicted $y$};
\node[anchor=west,inner sep=0pt] at (2.78,0.42) {\tikz\draw[inf](0,0)--(0.3,0);};
\node[tlab,anchor=west] at (3.20,0.42) {inferred edge};
\node[ev,fill=blue!50,scale=0.8] at (0.20,0.17) {};
\node[tlab,anchor=west] at (0.38,0.17) {evaluated};
\node[anchor=west,inner sep=0pt] at (2.78,0.17) {\tikz\draw[halo](0,0) ellipse (0.16 and 0.1);};
\node[tlab,anchor=west] at (3.20,0.17) {uncertainty};
\end{scope}
\begin{scope}[shift={(10.0,0)}]
\draw[panel] (0,0) rectangle (5.2,4.2);
\node[title] at (0,4.2) {(c) Edge query and node evaluation};
\node[tlab,anchor=north,font=\tiny\bfseries,text=black!70] at (1.05,3.72) {before};
\node[val,fill=blue!45] (Lb) at (0.4,2.3) {};
\node[val,fill=blue!78] (Ab) at (0.95,3.1) {}; \node[tlab,anchor=south,inner sep=0.5pt] at (0.95,3.18) {$\mathrm{A}$};
\draw[halo] (1.5,2.45) ellipse (0.4 and 0.26);
\node[val,fill=blue!70] (Bb) at (1.5,2.45) {}; \node[tlab,anchor=south west,inner sep=0.5pt] at (1.78,2.64) {$\mathrm{B}$};
\draw[unobs] (Lb)--(Bb) node[tlab,midway,below=-1pt,text=black!40] {?};
\node[tlab,anchor=east] at (0.55,1.4) {$\mathrm{A}$}; \draw[sbar] (0.6,1.4)--(1.62,1.4);
\node[tlab,anchor=east] at (0.55,1.05) {$\mathrm{B}$}; \draw[sbar] (0.6,1.05)--(1.5,1.05);
\draw[qarr] (2.05,2.2) -- (2.5,2.2);
\node[tlab,anchor=north,text=teal!70!black] at (2.28,2.12) {edge query};
\node[tlab,anchor=north,font=\tiny\bfseries,text=black!70] at (3.5,3.72) {after};
\node[val,fill=blue!45] (La) at (2.85,2.3) {};
\node[val,fill=blue!78] (Aa) at (3.4,3.1) {}; \node[tlab,anchor=south,inner sep=0.5pt] at (3.4,3.18) {$\mathrm{A}$};
\draw[halo] (3.95,2.45) ellipse (0.23 and 0.16);
\node[val,fill=blue!88] (Ba) at (3.95,2.45) {}; \node[tlab,anchor=south west,inner sep=0.5pt] at (4.23,2.64) {$\mathrm{B}$};
\draw[obs] (La)--(Ba) node[tlab,midway,above=-1pt,text=black!65] {0.7};
\node[tlab,anchor=east] at (3.0,1.4) {$\mathrm{A}$}; \draw[sbar] (3.05,1.4)--(4.07,1.4);
\node[tlab,anchor=east] at (3.0,1.05) {$\mathrm{B}$}; \draw[sbar] (3.05,1.05)--(4.25,1.05);
\node[nextnode,scale=0.7] at (4.38,1.05) {};
\node[tlab,anchor=north,text=red!60!black] at (3.7,0.9) {$v^{(t+1)}=\mathrm{B}$};
\node[tlab,anchor=south west,text=black!60] at (0.29,1.62) {Acquisition scores};
\node[tlab,anchor=south west,text=black!60] at (2.74,1.62) {Updated acquisition score};
\end{scope}
\draw[bigarr] (4.55,2.1)--(4.85,2.1); \node[tlab,anchor=south,text=black!65] at (4.7,2.14) {infer};
\draw[bigarr] (9.82,2.1)--(9.98,2.1); \node[tlab,anchor=south,text=black!65] at (9.9,2.14) {query};
\draw[bigarr,rounded corners=4pt] (12.6,-0.05) -- (12.6,-0.42) -- (2.25,-0.42) -- (2.25,-0.05);
\node[lab,anchor=north,fill=white,inner sep=1.5pt] at (7.425,-0.4) {Function evaluation $y(v^{(t+1)})$ and the new edge observations enter the next round};
\end{tikzpicture}}
\caption{{Global BO on a partially observed graph. (a) Two kinds of observation with two budgets: a few expensive values of the objective $y$, and entries of the adjacency matrix revealed by edge queries.  (b) A graph-valid low-rank surrogate fitted to the revealed entries places every node in one global geometry and gives every node a posterior and an acquisition score. A node that no query has touched has an uncertain feature and hence an uncertain score. (c) Structure is queried where it can change the next decision. An edge query against a landmark sharpens the feature of candidate $\mathrm B$, its score moves above that of $\mathrm A$, and $\mathrm B$ is evaluated next. The new value and the new entries enter the next round.}}
\label{fig:method-overview}
\end{figure}

\section{Methodology}
\label{sec:method}

We propose an iterative framework that alternates between (i) inference of a global surrogate geometry from the revealed entries and (ii) GP-based BO over the nodes on the induced geometry. At iteration $t$ the method holds the evaluated set $\mathcal V^{(t)}$ and the revealed set $\Omega^{(t)}$, and one round is the pipeline
\begin{align*}
\Omega^{(t)}
\rightarrow
\tilde{A}^{(t)}
\rightarrow
\tilde{L}_\zeta^{(t)}
\rightarrow
\tilde{K}^{(t)}
\rightarrow
{\text{GP posterior}}
\rightarrow
v^{(t+1)}
\rightarrow
\Omega^{(t+1)}.
\end{align*}
The revealed entries are fitted by a surrogate adjacency $\tilde A^{(t)}$ inside a graph-valid low-rank class, whose regularized normalized Laplacian $\tilde L_\zeta^{(t)}$ has a spectrum that is computed exactly in closed form. A spectral filter of this Laplacian is the kernel $\tilde K^{(t)}$ of a GP over all nodes, the next node $v^{(t+1)}$ maximizes an acquisition function of the posterior, and the edge queries of the round are chosen by their value for that decision, which closes the loop from the kernel back to $\Omega^{(t+1)}$. We query one node value per iteration, and the framework extends to batches.

The key insight is the following. As the surrogate is a valid rank-$d$ graph, every spectral kernel of its Laplacian is a white floor plus a rank-$d$ term, so the GP is a Bayesian linear model in $d$-dimensional node features with a coefficient shared by all nodes. Function evaluations learn the shared coefficient, edge queries learn the features, and the value of an edge query is the improvement of the next decision that its answer brings. Section~\ref{Subsec:graph surrogate} learns the geometry from the revealed entries. Section~\ref{Subsec:global BO} runs BO on the geometry. Section~\ref{Subsec:edge design} states what each revealed entry determines about the features and turns the value of a query into the edge-query rule of each query type, and Section~\ref{sec:theory} states what estimated features cost one decision, when the revealed entries identify the geometry, and sums the cost over rounds.  Algorithm~\ref{Alg:global_bo} (Appendix~\ref{app:method}) summarizes one round.

\subsection{Surrogate Graph Learning from Partial Observations}
\label{Subsec:graph surrogate}

We fit the revealed entries within a \emph{graph-valid low-rank class}. Each node has a membership $\pi_i\in\Delta_d:=\{b\ge0:\mathbf1^\top b\le1\}$, and memberships interact through a symmetric core:
\begin{equation}
\label{Equ:class}
\mathcal C_d=\bigl\{\Pi C\Pi^\top:\ \pi_i\in\Delta_d,\ C=C^\top\in[0,1]^{d\times d}\bigr\}.
\end{equation}
Every member is symmetric, has entries in $[0,1]$, and has rank at most $d$. Thus every member is a valid weighted graph by construction. The parameter $d$ controls the surrogate's resolution, without requiring the unknown graph itself to have rank $d$. Given revealed pairs $\Omega$, we solve
\begin{equation}
\label{Opt:fit}
\min_{\pi_i\in\Delta_d,\ C=C^\top\in[0,1]^{d\times d}}
\bigl\|\mathcal P_\Omega(\Pi C\Pi^\top-A)\bigr\|_F^2.
\end{equation}
We use projected alternating least squares with spectral initialization and warm starts (Appendix~\ref{app:coldstart}). Only the revealed entries enter the fit, and self-pairs never do.

\paragraph{Exact spectral geometry.}
For the fitted $\tilde A$, let $\tilde D=\operatorname{diag}(\tilde A\mathbf1)$ and $\tilde L_\zeta=I-(\tilde D+\zeta I)^{-1/2}\tilde A(\tilde D+\zeta I)^{-1/2}$, where $\zeta>0$ regularizes low degrees~\citep{chaudhuri2012spectral,qin2013regularized}. The normalized adjacency has rank at most $d$, so $\tilde L_\zeta$ has at most $d$ eigenvalues different from $1$. A thin QR factorization of the normalized memberships reduces the spectral solve to a $d\times d$ eigendecomposition, at total cost $\mathcal O(nd^2+d^3)$ (Lemma~\ref{lem:exact-solve}). This computes the surrogate's spectrum exactly without forming an $n\times n$ matrix. Write its nontrivial eigenpairs as $(\tilde\lambda_i,\tilde u_i)$.

\subsection{Global Bayesian Optimization on the Surrogate}
\label{Subsec:global BO}

The eigenvalue $1$ on the orthogonal complement provides a natural bulk level for a spectral filter~\citep{smola2003kernels}. With a positive decreasing response $r$, we retain the positive contrasts $r(\tilde\lambda_i)-r(1)$ and give the structured and bulk components separate scales:
\begin{equation}
\label{Equ:kernel2}
\tilde K=\sigma_b^2I+\sigma_s^2\tilde U_+D_+\tilde U_+^\top,
\qquad D_+=\operatorname{diag}\bigl(r(\tilde\lambda_i)-r(1)\bigr)_{\tilde\lambda_i<1}.
\end{equation}
Here $\tilde U_+$ contains the $d_+\le d$ retained modes. This \emph{structured-plus-bulk kernel} is positive definite for $\sigma_b^2>0$. Retaining positive contrasts permits independent Gaussian coefficients. A one-scale filter ties their variances to the bulk level, whereas separate scales let the data tune the balance between cross-node generalization and node-specific variation. We use the resolvent $r(\lambda)=(\beta+\lambda)^{-1}$ and select $(\sigma_b^2,\sigma_s^2,\beta,\zeta)$ by marginal likelihood. The exact one-scale alternative is given in Appendix~\ref{app:method}.

Let $\phi_v$ be the transposed $v$-th row of $\tilde U_+$ and $D=\sigma_s^2D_+$. Conditional on the fitted geometry, the kernel is equivalent to $y(v)=\phi_v^\top\theta+\varepsilon_v$, with $\theta\sim\mathcal N(0,D)$ and independent $\varepsilon_v\sim\mathcal N(0,\sigma_b^2)$. The residual is drawn once per node, and observations of $y$ are exact. If $w_t,\Xi_t$ are the posterior mean and covariance of $\theta$, then for an unevaluated node
\begin{equation}
\label{Equ:prediction}
\mu_t(v)=\phi_v^\top w_t,\qquad s_t^2(v)=\sigma_b^2+\phi_v^\top\Xi_t\phi_v.
\end{equation}
The coefficient update uses only a $d_+\times d_+$ system (Lemma~\ref{lem:reduction}). This representation separates the two learning tasks: evaluations update the shared coefficient, and structural queries refine node features. We select the next node using the plug-in GP-UCB score on the current estimated features,
\begin{equation}
\label{Equ:ucb-struct}
U_t(v)=\mu_t(v)+\kappa s_t(v),\qquad
v^{(t+1)}=\arg\max_{v\notin\mathcal V^{(t)}}U_t(v).
\end{equation}
Here $\kappa>0$ controls exploration.

\subsection{Decision-Aware Structure Acquisition}
\label{Subsec:edge design}

\paragraph{From entries to feature uncertainty.}
Conditional on the core and a reference node's membership, an entry in the surrogate model is a linear measurement, $\tilde A_{vu}=g_u^\top\pi_v$ with $g_u=C\pi_u$. We call these reference nodes \emph{landmarks}. They are the revealed rows under row access and the initial block's nodes under pair access. Their memberships are fitted from the available entries. In the local acquisition model, we hold the fitted global quantities fixed and use a Gaussian working covariance $\Sigma_v$ to track the unmeasured membership directions. A node is \emph{pinned} when its measurement directions span $\mathbb R^d$, making $\Sigma_v=0$ conditional on that geometry. The feature map is differentiable in $\pi_v$, giving $\Sigma_v^\phi=J_v\Sigma_vJ_v^\top$, where $J_v=\partial\phi_v/\partial\pi_v$. The map and conditioning formulas are in Appendix~\ref{app:method}.

With feature mean $\hat\phi_v$ and covariance $\Sigma_v^\phi$ independent of the coefficient posterior in this working model, total variance separates three contributions:
\begin{equation}
\label{Equ:three-term}
\tilde s_t^2(v)=\underbrace{\sigma_b^2}_{\text{bulk}}
+\underbrace{\hat\phi_v^\top\Xi_t\hat\phi_v}_{\text{coefficients}}
+\underbrace{w_t^\top\Sigma_v^\phi w_t+\operatorname{tr}(\Sigma_v^\phi\Xi_t)}_{\text{features}}.
\end{equation}
Function observations inform the second term, while structural measurements reduce the third under the fixed local model. The latter weights feature uncertainty by its relevance to the objective. This uncertainty is conditional on the fitted global quantities. We use it to value structural queries, retain the plug-in node score~\eqref{Equ:ucb-struct}, and account for the actual structural score error separately in Section~\ref{sec:theory}. Because evaluated-node features affect all predictions through the shared coefficient posterior, both query rules prioritize them within a pinning allowance $\rho_{\mathrm{pin}}$. Until pinned, their coefficient updates use the moment adjustment in Eq.~\eqref{Equ:eiv}.

\paragraph{Pair queries.}
A useful pair should reveal a direction that matters to a competitive node's score. Holding the geometry and coefficient state fixed, let $c_v=J_v^\top(w_t+\kappa\Xi_t\hat\phi_v/s_t(v))$ be its membership sensitivity. A query along $g$ changes the linearized score mean by a zero-mean Gaussian increment with standard deviation
\begin{equation}
\label{Equ:delta}
\delta(v,g)=\frac{|c_v^\top\Sigma_vg|}{\sqrt{g^\top\Sigma_vg}},\qquad g^\top\Sigma_vg>0.
\end{equation}
Let $\mathrm{gap}_t(v)=|U_t(v)-\max_{w\ne v}U_t(w)|$, with the maximum over other unevaluated nodes. The one-query knowledge gradient~\citep{frazier2008knowledge} is $\delta f(-\mathrm{gap}_t(v)/\delta)$, where $f(z)=z\Phi(z)+\varphi(z)$ and $\Phi,\varphi$ are the standard normal distribution and density. It is the expected reduction in selection loss for the local linearized scores (Proposition~\ref{prop:kg}). It increases with the informative score variation $\delta$ and decreases with the competitive gap.

To account for measurements whose value emerges jointly, we use a KG($*$) lookahead~\citep{frazier2010paradoxes}:
\begin{equation}
\label{Equ:kg}
\mathrm{KG}_t^*(v)=\max_{1\le m\le m_v}
\frac{\delta_m(v)}{m}f\!\left(-\frac{\mathrm{gap}_t(v)}{\delta_m(v)}\right),
\quad \delta_m(v)^2=c_v^\top(\Sigma_v-\Sigma_v^{(m)})c_v.
\end{equation}
Here $\Sigma_v^{(m)}$ follows $m$ greedy best-direction queries with $c_v$ fixed, and $m_v$ is the number of available independent directions, at most $d-\operatorname{rank}G_v$, where $G_v$ stacks the measured directions. Covariance updates do not depend on the answers, so plans can be scored before querying. Among unevaluated nodes with an informative unobserved landmark pair, we choose the largest score, execute its first query, and replan after the answer. Zero-variance plans have value zero. Algorithm~\ref{Alg:pair} gives the allocation details.

\paragraph{Row queries.}
A revealed row measures every other node along a common direction, so all unrevealed nodes share the same membership covariance, although its effect on their features and scores is node dependent (Proposition~\ref{prop:shared}). Before the revealed rows span the membership space a row adds one direction to every unrevealed node at once, and afterwards it adds nothing to the working covariance, which leaves a row rule little to allocate beyond the node it pins. We therefore reveal the rows of evaluated nodes first, whose feature errors enter every prediction through the shared coefficient, and sample the remaining rows uniformly without replacement (Algorithm~\ref{Alg:structure}). Section~\ref{sec:exp-q3} compares this rule with revealing the row of the current leader, which targets the pinned node, and with local rows.

\subsection{Theoretical Guarantees}
\label{sec:theory}

We analyze how estimated geometry affects optimization relative to a fixed reference surrogate: a full-information minimizer of~\eqref{Opt:fit}, with fixed parameters in~\eqref{Equ:kernel2}. Let $(\mu_t^\circ,s_t^\circ)$ and $U_t^\circ=\mu_t^\circ+\kappa s_t^\circ$ be its posterior and score on the actual evaluated set. Define $\Delta_t(v)=|U_t(v)-U_t^\circ(v)|$, which includes errors in both the candidate features and the shared coefficient estimate.

\begin{assumption}[Confidence for the reference model]
\label{ass:confidence}
For every $t<T$ and unevaluated $v$, $|y(v)-\mu_t^\circ(v)|\le\kappa s_t^\circ(v)$.
\end{assumption}
For a draw from the fixed reference GP this holds with probability at least $1-\delta_0$ for $\kappa^2=2\log(2nT/\delta_0)$, whatever the evaluation and query rules, since edge queries are functions of the history and of the fixed matrix $A$~\citep{srinivas2010gaussian}. Appendix~\ref{app:derivations}(H) also gives a deterministic sufficient condition.

When the optimum is unevaluated, maximizing $U_t$ incurs the per-round cost
\begin{equation}
\label{Equ:perround}
y(v^*)-y(v^{(t+1)})\le2\kappa s_t^\circ(v^{(t+1)})+\Delta_t(v^{(t+1)})+\Delta_t(v^*).
\end{equation}
Thus the geometry is charged through score errors at the selected node and the optimum. These errors include the influence of evaluated features on all predictions. With the reference bulk variance and coefficient state, they reduce to feature errors via $\Delta_t(v)\le\Lambda_t\|\hat\phi_v-\phi_v\|_2$, with $\Lambda_t=\|w_t\|_2+\kappa\|\Xi_t\|_2^{1/2}$ in a common feature basis (Lemma~\ref{lem:optimism}). This links the local score sensitivity used for acquisition to the decision error in the analysis.

\begin{theorem}[Simple regret with estimated geometry]
\label{thm:regret}
Under Assumption~\ref{ass:confidence}, let $v_{\max}=\max_v\phi_v^\top D\phi_v$ for the fixed reference kernel. Then
\[
r_T\le2\kappa\sigma_b+2\kappa\sqrt{\frac{c_1\gamma_T}{T}}
+\frac1T\sum_{t<T}\bigl(\Delta_t(v^{(t+1)})+\Delta_t(v^*)\bigr),
\]
where $c_1=2\max(\sigma_b^2,v_{\max})$ and $\gamma_T=d_+\log\bigl(1+(T+N_0)v_{\max}/(d_+\sigma_b^2)\bigr)$, with $\gamma_T=0$ if $d_+=0$.
\end{theorem}
The terms separate node-specific variation, learning in $d_+$ structured directions, and the cost of estimating the geometry. The bound applies to both query types. Its proof and a version for subsets of rounds are in Appendix~\ref{app:derivations}(H).

\paragraph{When structure is identified.}
If $A_{\mathrm{off}}$ has a completion in $\mathcal C_d$, $C$ is nonsingular, every $d$ memberships are independent, and $n\ge2d+1$, then $d+1$ distinct rows identify the surrogate, whereas $d$ rows need not suffice (Proposition~\ref{prop:coverage}). With the reference kernel parameters, a zero-residual fit, and exact coefficient conditioning thereafter, $\Delta_t=0$ after identification. Applying the bound to those rounds gives $r_T\le2\kappa\sigma_b+2\kappa\sqrt{c_1\gamma_T/(T-t_{\mathrm{id}})}$, where $t_{\mathrm{id}}$ counts decisions before identification (Corollary~\ref{cor:rows}). Pair queries instead control node-specific measurement sets. Their identification conditions and feature-error specialization, together with the local estimation bound outside the class, are in Appendix~\ref{app:recovery}.

\section{Experiments}
\label{sec:experiments}
\begingroup
We organize the experiment results around three questions. \textbf{(Q1)} When objective evaluations and structural queries are both budgeted, does a global surrogate learned from the revealed entries find high-value nodes that structure-free search and local graph BO do not? \textbf{(Q2)} How much of the remaining gap to a fully observed graph is due to the rank-$d$ representation, and how much to estimating it from partial observations? \textbf{(Q3)} When does decision-aware acquisition of structure improve on uninformed queries? Protocols, complete sweeps, and controls are in Appendix~\ref{app:exp}.

\subsection{Experimental Setup}
\label{sec:exp-setup}
\begin{figure}[t]
\centering
\includegraphics[width=0.55\linewidth]{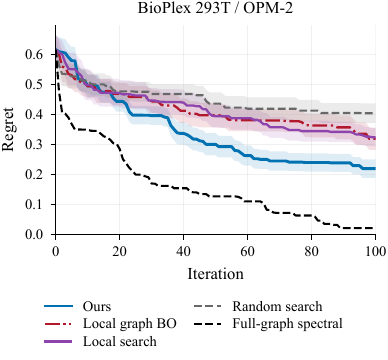}
\caption{Row queries on BioPlex 293T with the OPM-2 essentiality objective: mean normalized simple regret with pointwise $\pm1$ SEM bands over 20 paired seeds, $Q_R=10$ rows per round. All non-oracle arms come from the campaign of Table~\ref{tab:results}, and local graph BO~\citep{wan2023bayesian} receives the same row allowance. The dashed full-graph spectral reference ($d=16$) sees the entire graph. The dense and ESOL trajectories of the same campaign are in Figure~\ref{fig:exp-dense-rows}.}
\label{fig:rows-main}
\end{figure}
\textbf{Tasks.} \emph{In-class}: dense affinities on $n=1{,}000$ nodes drawn from the class at rank $16$, with two objectives, a draw from the class-kernel prior and a latent objective defined without reference to the kernel. \emph{Real}: ESOL, $1{,}117$ molecules with Tanimoto affinities and measured solubility~\citep{delaney2004esol}, BioPlex~3.0 protein interaction networks assayed in HEK293T ($n=13{,}167$) and HCT116 ($n=9{,}609$) cells with DepMap gene-essentiality objectives~\citep{huttlin2021dual,tsherniak2017defining}, the negated CRISPR gene effect of each protein's gene averaged over all screened cell lines (mean essentiality) or taken in one line, the colon adenocarcinoma line C2BBe1, the myeloma line OPM-2, and HCT116 itself for the HCT116 network, and Chameleon ($n=1{,}223$), a Wikipedia page graph with log traffic as the objective~\citep{rozemberczki2021multi}. Sparse DCSBM and latent-position graphs are in Appendix~\ref{app:exp-additional}.
\textbf{Protocol.} Every method starts from the same $N_0$ evaluated nodes within each task and seed. Our row and pair methods receive the rows or the pair block among those nodes, then alternate one evaluation with an allowance of $Q_R$ rows or $Q_E$ pairs per round for $T$ rounds, at the budgets of Table~\ref{tab:exp-protocols}.
\textbf{Methods.} \emph{Ours} uses the structured-plus-bulk kernel with $d=16$ in the row studies, chosen on separate design seeds and held fixed across tasks, and $d=32$ in the pair study, with pin-then-uniform rows or KG$^*$ pairs and hyperparameters by marginal likelihood on the revealed data. We compare with random node search, which buys no structure, local search with restarts over the revealed neighborhoods, and local graph BO~\citep{wan2023bayesian}, which explores the revealed neighborhood with a local kernel and receives the same structural allowance as Ours under either query type. On the dense and ESOL affinities its neighborhoods come from a precomputed $k$NN view of the full graph (Appendix~\ref{app:exp-baselines}). Full-graph spectral BO on the true Laplacian sees the entire graph and is the oracle reference.
\textbf{Metric.} Normalized simple regret $R_t=\bigl(y^*-\max_{v\in\mathcal V^{(t)}}y(v)\bigr)/\bigl(y^*-\operatorname{median}_v y(v)\bigr)$ after $t$ evaluations, mean $\pm$ SEM over 20 seeds that share the initial nodes.

\subsection{Optimization under Budgeted Structural Access}
\label{sec:exp-q1}
\begin{figure}[t]
\centering
\includegraphics[width=0.62\linewidth]{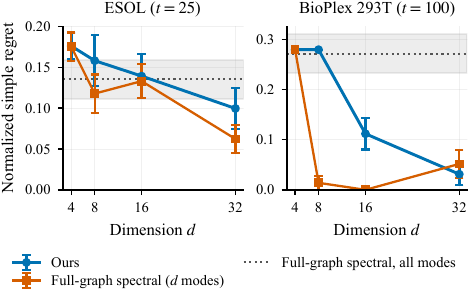}
\caption{Representation under full graph access, mean $\pm1$ SEM over 20 seeds: the class fitted to the entire graph against the $d$ smoothest modes of the true Laplacian, with the all-mode kernel, tuned on its own grid, as a reference. BioPlex at $t=100$, ESOL at $t=25$, before the regret floor is reached.}
\label{fig:exp-resolution}
\end{figure}
Table~\ref{tab:results} and Figure~\ref{fig:rows-main} answer Q1 with one complete, paired-seed row-query campaign, in which roughly one row in ten is revealed on the dense, ESOL, and BioPlex instances (Table~\ref{tab:exp-protocols}). Ours has the lowest mean final regret among the four non-oracle methods on all ten objectives, including the dense latent objective, which is not drawn from the kernel. On the dense in-class objective the fit to the revealed rows is within seed variability of the class fitted to the entire graph and below the full-graph spectral reference at the same $d$ (Table~\ref{tab:exp-structure-references}), so the revealed rows already carry the geometry. Ours closes most of the distance from random search to the full-graph spectral reference on the dense, Chameleon, and HCT116 objectives, and between a third and two thirds of it on ESOL and the 293T objectives, where Section~\ref{sec:exp-q2} decomposes the remaining gap. On BioPlex the advantage appears within the first twenty evaluations and persists (Figure~\ref{fig:rows-main}), and Figure~\ref{fig:exp-row-budgets} places the $1{,}000$-row column inside the full row-budget sweep.

\begin{table}[t]
\centering\footnotesize
\setlength{\tabcolsep}{3pt}
\caption{Row queries: final normalized simple regret, mean $\pm$ SEM over 20 paired seeds, at the budgets of Table~\ref{tab:exp-protocols}. Bold marks the lowest mean among the four non-oracle methods. All methods receive the same structural allowance per round, with the neighborhood construction of local graph BO~\citep{wan2023bayesian} described in Appendix~\ref{app:exp-protocols}. The full-graph spectral reference ($d=16$) observes the graph before BO.}
\label{tab:results}
\begin{tabular}{lcccc|c}
\toprule
Instance / objective & Ours & Random & Local search & Local graph BO & \shortstack{Full graph\\spectral}\\\midrule
Dense / class prior & $\mathbf{.008\pm.008}$ & $.134\pm.020$ & $.241\pm.058$ & $.330\pm.060$ & $.043\pm.026$\\
Dense / latent & $\mathbf{.158\pm.013}$ & $.201\pm.027$ & $.196\pm.024$ & $.187\pm.023$ & $.170\pm.026$\\
\midrule
ESOL / solubility & $\mathbf{.061\pm.014}$ & $.147\pm.018$ & $.068\pm.013$ & $.095\pm.015$ & $.005\pm.005$\\
Chameleon / log traffic & $\mathbf{.079\pm.005}$ & $.097\pm.012$ & $.108\pm.024$ & $.088\pm.023$ & $.077\pm.001$\\
293T / mean essentiality & $\mathbf{.304\pm.042}$ & $.468\pm.032$ & $.376\pm.035$ & $.333\pm.045$ & $.000\pm.000$\\
293T / C2BBe1 & $\mathbf{.226\pm.038}$ & $.513\pm.034$ & $.446\pm.032$ & $.437\pm.034$ & $.000\pm.000$\\
293T / OPM-2 & $\mathbf{.219\pm.031}$ & $.404\pm.033$ & $.323\pm.031$ & $.319\pm.037$ & $.021\pm.014$\\
HCT116 / mean essentiality & $\mathbf{.323\pm.023}$ & $.408\pm.025$ & $.339\pm.010$ & $.341\pm.030$ & $.330\pm.003$\\
HCT116 / C2BBe1 & $\mathbf{.349\pm.010}$ & $.456\pm.031$ & $.363\pm.012$ & $.400\pm.019$ & $.330\pm.001$\\
HCT116 / own line (HCT116) & $\mathbf{.359\pm.021}$ & $.476\pm.021$ & $.392\pm.019$ & $.389\pm.033$ & $.298\pm.003$\\
\bottomrule
\end{tabular}
\end{table}

\subsection{Surrogate Representation and Resolution}
\label{sec:exp-q2}
Full graph access separates the two parts of the gap to the oracle. The estimation gap of a task is the regret of Ours on the revealed rows minus that of the same class fitted to the entire graph, and the representation gap is the latter minus the full-graph spectral reference at the same $d$ (Tables~\ref{tab:results} and~\ref{tab:exp-structure-references}, Appendix~\ref{app:exp-additional}). On the dense instance and Chameleon the partial fit is at or below the full fit, so what remains is representation. On ESOL and on 293T with mean essentiality estimation dominates, on the other two 293T objectives the two gaps are comparable, and on HCT116 both are small. Figure~\ref{fig:exp-resolution} sweeps the resolution $d$ under full graph access. On the sparse binary BioPlex graph the eigenbasis reaches its floor at $d=8$ while the class needs $d=32$, so the class pays for not having the eigenbasis with a higher resolution, on ESOL the class improves across the sweep, and on DCSBM both arms improve up to $d=16$ (Figure~\ref{fig:exp-dcsbm-resolution}). At $d=32$ every truncated arm is below the all-mode kernel on all three instances, and under the row-study protocol the all-mode kernel is also above the class fitted to the entire graph on the dense and ESOL instances (Table~\ref{tab:exp-structure-references}), so at these resolutions truncation costs no regret relative to the untruncated kernel.

\subsection{Value of Decision-Aware Structure Acquisition}
\label{sec:exp-q3}
\begin{figure}[t]
\centering
\includegraphics[width=0.62\linewidth]{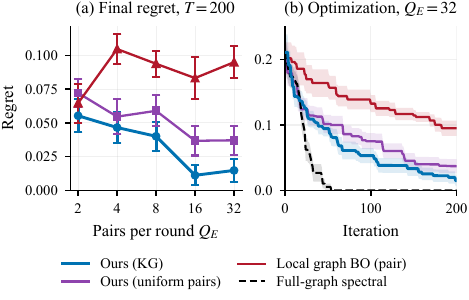}
\caption{Pair queries on ESOL, $d=N_0=32$, $T=200$, 20 shared seeds. (a) Final normalized regret against the pairs acquired per round, $\pm1$ SEM. (b) Regret trajectories at $Q_E=32$ with $\pm1$ SEM bands. The dashed full-graph spectral reference sees the entire graph. All five budgets are in Figure~\ref{fig:exp-pair-round}.}
\label{fig:exp-acquisition}
\end{figure}
Under pair queries each revealed entry informs one node, and Section~\ref{Subsec:edge design} prices it by its expected effect on the next decision. Figure~\ref{fig:exp-acquisition} tests this on ESOL with $d=N_0=32$ and $T=200$. All methods start from the same evaluated nodes and the pair block among them, and acquire $Q_E\in\{2,4,8,16,32\}$ new pairs before every evaluation. Ours (KG) has the lowest final regret at every budget. Its gap to local graph BO, adapted to the pair oracle, opens immediately and is significant at every $Q_E\ge4$ after adjusting for the five budgets. Its gap to uniform pairs, which share the surrogate and the kernel, opens after about fifty evaluations, has the same sign at every budget, and is largest at the two largest budgets, where the regret of Ours (KG) is less than half that of uniform pairs (Table~\ref{tab:exp-pair-ci}). The difference between the two Ours arms is the acquisition-and-conditioning procedure of Section~\ref{Subsec:edge design}, which spends each query on a candidate whose feature is undetermined and whose score is close to the top.

Under row queries, neither UCB-leader rows nor local BFS rows improves on pin-then-uniform on ESOL consistently, with differences comparable to the SEM (Table~\ref{tab:exp-row-rules}). Together with the shared measurement pattern of Proposition~\ref{prop:shared}, these results support the simple coverage-based default. The pair study evaluates the benefit of targeting structure at competitive candidates.

\endgroup

\section{Conclusion}
We studied Bayesian optimization on graphs whose affinities are revealed only through budgeted edge queries, and replaced local exploration of the revealed subgraph by a global geometry inferred from it. A graph-valid rank-$d$ surrogate with an exactly computable spectrum makes the GP a linear model in node features, in which evaluations learn the shared coefficient and edge queries learn the features, so that each edge query is priced by its expected effect on the next decision. The analysis charges the estimated geometry at two nodes, $d+1$ rows identify an in-class surrogate, and structural access enters the regret bound only through the identification time. Across protein interaction networks, a molecular similarity graph, a Wikipedia page graph, and synthetic graphs, the method attains the lowest final regret among the non-oracle methods on all ten row-query objectives, the decision-aware pair rule improves on uniform pairs at every budget, and full-information controls separate the representation gap from the estimation gap.

\bibliography{Ref_BOgraph}
\bibliographystyle{plainnat}

\appendix
\makeatletter
\@addtoreset{lemma}{section}\@addtoreset{proposition}{section}\@addtoreset{corollary}{section}\@addtoreset{remark}{section}\@addtoreset{assumption}{section}
\makeatother
\renewcommand{\thelemma}{\thesection.\arabic{lemma}}
\renewcommand{\theproposition}{\thesection.\arabic{proposition}}
\renewcommand{\thecorollary}{\thesection.\arabic{corollary}}
\renewcommand{\theremark}{\thesection.\arabic{remark}}
\renewcommand{\theassumption}{\thesection.\arabic{assumption}}
\providecommand{\theHlemma}{}\renewcommand{\theHlemma}{\thesection.\arabic{lemma}}
\providecommand{\theHproposition}{}\renewcommand{\theHproposition}{\thesection.\arabic{proposition}}
\providecommand{\theHcorollary}{}\renewcommand{\theHcorollary}{\thesection.\arabic{corollary}}
\providecommand{\theHremark}{}\renewcommand{\theHremark}{\thesection.\arabic{remark}}
\providecommand{\theHassumption}{}\renewcommand{\theHassumption}{\thesection.\arabic{assumption}}

\section{Identification and Estimation of the Surrogate}
\label{app:recovery}
We collect the identification and estimation results used in Section~\ref{sec:theory}, together with the row and pair specializations of its regret bound.

\begin{corollary}[Row queries in the class]
\label{cor:rows}
Suppose $A_{\mathrm{off}}$ has a completion in $\mathcal C_d$. Let $t_{\mathrm{id}}$ count decisions before the observed rows identify that completion, and suppose the solver returns a zero-residual fit thereafter. Under Theorem~\ref{thm:regret}, using the reference kernel parameters and exact coefficient conditioning after identification, for $T>t_{\mathrm{id}}$,
\[
r_T\le2\kappa\sigma_b+2\kappa\sqrt{c_1\gamma_T/(T-t_{\mathrm{id}})}.
\]
If $C$ is nonsingular, memberships are in general position, $n\ge2d+1$, and $N_0$ initial rows are followed by $Q_R\ge1$ distinct rows before each decision, then
$t_{\mathrm{id}}\le(\lceil(d+1-N_0)_+/Q_R\rceil-1)_+$.
For interior representatives this threshold is also necessary in general.
\end{corollary}
The proof is in Appendix~\ref{app:derivations}(H). The bound concerns identification from observations and recovery by the solver as separate conditions.

\begin{assumption}[Split smoothness]
\label{ass:split}
Let $K_s=\sigma_s^2\,\tilde U_+ D_+ \tilde U_+^\top$ be the structured part of the kernel of Eq.~\eqref{Equ:kernel2} built from the reference surrogate. There are known $B_s>0$ and $\eta\ge0$ such that $y=y_s+y_b$ with $\|y_s\|_{\mathcal H(K_s)}\le B_s$ and $\|y_b\|_\infty\le\eta$.
\end{assumption}
This condition separates variation on the structured modes from a bounded node-specific residual. Appendix~\ref{app:derivations}(H) gives the resulting deterministic confidence condition.

\begin{remark}[Node-specific variation under a finite evaluation budget]
\label{rem:floor}
In Assumption~\ref{ass:split}, take $y_s=0$ and $y_b=\eta\mathbf1_{v_0}$, where $v_0$ is uniform on $\mathcal V$ independently of $A$. Until $v_0$ is evaluated, the observations match those of the zero objective. With $N_0+T<n$ evaluations, the probability of finding it is at most $(N_0+T)/n$, regardless of structural queries. Hence expected simple regret is at least $(1-(N_0+T)/n)\eta$, and some fixed spike location has at least this expected regret over the algorithm's randomness. The bound holds for every structural budget and every adaptive query rule, since edge queries reveal $A$ and not $y$.
\end{remark}

Under pair queries the errors are node specific, and Theorem~\ref{thm:regret} specializes as follows.

\begin{corollary}[Pair queries]
\label{cor:pairs}
Under the conditions of Theorem~\ref{thm:regret}, if at every decision the algorithm uses the reference kernel parameters and the evaluated nodes are pinned with exact features, then every pair-query rule satisfies
\[
r_T\ \le\ 2\kappa\,\sigma_b\ +\ 2\kappa\sqrt{\frac{c_1\,\gamma_T}{T}}\ +\ \frac{1}{T}\sum_{t<T}\Lambda_t\bigl(e_t(v^{(t+1)})+e_t(v^*)\bigr),
\]
with $e_t(v)=\|\hat\phi_v-\phi_v\|_2$ in a common feature basis at decision time. If moreover $A_{\mathrm{off}}\in\mathcal C_d$, the landmark skeleton is identified with exact global quantities $(R,V,h)$, and the spanning and pinning-allowance conditions of Lemma~\ref{lem:leader} hold, Algorithm~\ref{Alg:pair} makes both errors vanish in every round with $Q_E\ge d(m_t+1)$, where $m_t$ is the number of unpinned unevaluated nodes at the start of the round.
\end{corollary}
The proof is in Appendix~\ref{app:derivations}(L).

\paragraph{Identification.}
Two conventions. A set of membership vectors is in \emph{general position} if every $d$ of them are linearly independent. A representative $(\Pi,C)$ of a class member is \emph{interior} if every $\pi_i$ lies in the interior of $\Delta_d$ and $C$ is nonsingular with entries in $(0,1)$. Throughout, $M=\Pi C\Pi^\top$ denotes the surrogate including its diagonal, which is the object the kernel uses, and the revealed entries are entries of $M_{\mathrm{off}}=A_{\mathrm{off}}$.

\begin{proposition}[Identification]
\label{prop:coverage}
Let $A_{\mathrm{off}}=M_{\mathrm{off}}$ with $M=\Pi C\Pi^\top$, $\pi_i\in\Delta_d$, $C=C^\top$ nonsingular, and $\operatorname{rank}\Pi=d$.
(i) \emph{Count.} The set of such $M$ contains a nonempty open subset of the manifold of symmetric matrices of rank $d$, whose dimension is $nd-\binom d2$, so no set of fewer than $nd-\binom d2$ entries determines $M$ for every member of the class.
(ii) \emph{Rows.} Let $\mathcal R$ be the revealed rows and $r=|\mathcal R|$. If $\operatorname{rank}\Pi_{\mathcal R\setminus\{u\}}=d$ for every $u\in\mathcal R$ and $\operatorname{rank}\Pi_{\mathcal V\setminus\mathcal R}=d$, then $M$ is the only symmetric matrix of rank at most $d$ that agrees with $A$ on the revealed entries, and every zero-residual point of Problem~\eqref{Opt:fit} equals $M$. Both conditions hold when $r\ge d+1$, $n-r\ge d$, and the memberships are in general position, and for pure memberships $\pi_i=\theta_ie_{k(i)}$ with $\theta_i>0$ they hold exactly when every community has two revealed rows and one unrevealed node. Conversely, if $r\le d$, $\operatorname{rank}\Pi_{\mathcal R}=r$, and the representative is interior, the revealed rows do not determine $M$.
(iii) \emph{Pairs.} Let $\mathcal R$ be a set of $r$ nodes with all pairs among them revealed. If $r<r_0(d):=\lceil(2d+1+\sqrt{8d+1})/2\rceil$, the revealed block does not determine $M_{\mathcal R\mathcal R}$ for every member of the class. If $\mathcal R$ contains disjoint $\mathcal R_1,\mathcal R_2$ with $\operatorname{rank}\Pi_{\mathcal R_1\setminus\{u\}}=d$ for every $u\in\mathcal R_1$ and $\operatorname{rank}\Pi_{\mathcal R_2}=d$, which holds for $r\ge2d+1$ in general position, then the block determines $M_{\mathcal R\mathcal R}$. A node $v\notin\mathcal R$ whose entries against $R_v\subseteq\mathcal R$ are revealed with $\operatorname{rank}\Pi_{R_v}=d$ has $M_{v,\mathcal R}$ determined, and $M$ is determined once every node is so measured.
\end{proposition}
The proof is in Appendix~\ref{app:derivations}(G). The row design with $r=d+1$ reveals $(d+1)(n-1)-\binom{d+1}2$ distinct entries and the pair design $\binom r2+(n-r)d$, both of order $nd$, so both are order-optimal by (i). The count in (iii) is Ledermann's bound of factor analysis~\citep{ledermann1937rank}: $\binom r2\ge rd-\binom d2$ is the condition under which a symmetric $r\times r$ matrix of rank $d$ with unknown diagonal has no more free parameters than observed entries. For positive semidefinite low-rank parts, generic identifiability of the low-rank part from the off-diagonal entries is classical when Ledermann's bound is strictly satisfied~\citep{shapiro1985identifiability,bekker1997generic}. Our core $C$ is indefinite in general, and we use only the constructive condition $r\ge2d+1$. Completion from uniformly sampled entries needs of the order of $\mu_0nd\log n$ entries under an incoherence assumption with parameter $\mu_0$~\citep{candes2009exact}, whereas the landmark designs need no logarithmic factor and no incoherence, because every node is measured against the same landmarks and the problem decouples into $n$ small linear systems. The proof of (ii) is constructive and exhibits the estimator as a skeleton decomposition $M=M_{\cdot,\mathcal R}M_{\mathcal R\mathcal R}^{+}M_{\mathcal R,\cdot}$~\citep{mahoney2009cur}, once the $r$ diagonal entries of the landmark block are filled in from the column space of $M_{\mathcal R,\mathcal V\setminus\mathcal R}$.

\paragraph{Estimation error outside the class.}
Let $(\Pi^\star,C^\star)$ be the reference fit, a minimizer of Eq.~\eqref{Equ:bias}, with features $\phi^\star_v$, and let $(\hat\Pi,\hat C)$ be a fit to the revealed rows $\mathcal R$. Write $B^\star=\Pi^\star_{\mathcal R}C^\star$ and $\hat B=\hat\Pi_{\mathcal R}\hat C$ for the two $r\times d$ matrices of landmark directions, and $\rho_v=A_{\mathcal R,v}-B^\star\pi^\star_v$ for the residual of the reference fit on the revealed entries of an unrevealed node $v$, which is gauge invariant and vanishes in the class.

\begin{proposition}[Per-node estimation error under row queries]
\label{prop:node-error}
Let $\hat B$ have rank $d$ and let $\hat\pi_v$, $v\notin\mathcal R$, minimize Problem~\eqref{Opt:fit} over $\pi_v\in\Delta_d$ with $(\hat\Pi_{\mathcal R},\hat C)$ fixed, as every minimizer of Problem~\eqref{Opt:fit} does, since the objective is separable across unrevealed nodes. Then for every $v\notin\mathcal R$,
\[
\|\hat\pi_v-\pi^\star_v\|_2\ \le\ \frac{\|P_{\hat B}\,\rho_v\|_2+\|\hat B-B^\star\|_2}{\sigma_{\min}(\hat B)}\ \le\ \frac{\|\hat B^\top\rho_v\|_2}{\sigma_{\min}(\hat B)^2}+\frac{\|\hat B-B^\star\|_2}{\sigma_{\min}(\hat B)},
\]
where $P_{\hat B}$ is the orthogonal projector onto the column space of $\hat B$. With the reference feature map $\psi(\pi)=F\pi/\sqrt{\pi^\top h+\zeta}$ of Eq.~\eqref{Equ:feature} held fixed, $\psi$ is Lipschitz on $\Delta_d$ with $L_\phi\le\|F\|_2\zeta^{-1/2}(1+\|h\|_2/(2\zeta))$. Thus $\|\psi(\hat\pi_v)-\psi(\pi^\star_v)\|_2$ obeys the same bound multiplied by $L_\phi$. The statement holds for every gauge-equivalent representative $(\Pi^\star S,S^{-1}C^\star S^{-\top})$ of the reference fit whose memberships lie in $\Delta_d$.
\end{proposition}
\begin{proof}
Fix $v\notin\mathcal R$ and write $a=A_{\mathcal R,v}$. The terms of Problem~\eqref{Opt:fit} that involve $\pi_v$ are proportional to $f(\pi)=\tfrac12\|a-\hat B\pi\|_2^2$, so $\hat\pi_v\in\arg\min_{\pi\in\Delta_d}f(\pi)$. Since $\hat B$ has full column rank, $f(\pi)=\tfrac12\|\hat B(\pi-\tilde\pi)\|_2^2+\tfrac12\|(I-P_{\hat B})a\|_2^2$ with $\tilde\pi=\hat B^{+}a$, so $\hat\pi_v$ is the projection of $\tilde\pi$ onto the closed convex set $\Delta_d$ in the norm $\|x\|_{\hat B}=\|\hat Bx\|_2$. Metric projections onto convex sets are nonexpansive and fix the points of the set, so for every $\pi\in\Delta_d$, $\|\hat B(\hat\pi_v-\pi)\|_2\le\|\hat B(\tilde\pi-\pi)\|_2=\|P_{\hat B}(a-\hat B\pi)\|_2$. Take $\pi=\pi^\star_v$ and write $a-\hat B\pi^\star_v=\rho_v+(B^\star-\hat B)\pi^\star_v$. Since $\|\pi^\star_v\|_2\le\|\pi^\star_v\|_1\le1$ and $\|P_{\hat B}\|_2\le1$, we get $\|\hat B(\hat\pi_v-\pi^\star_v)\|_2\le\|P_{\hat B}\rho_v\|_2+\|\hat B-B^\star\|_2$, and dividing by $\sigma_{\min}(\hat B)$ gives the first bound. The second follows from $P_{\hat B}=\hat B(\hat B^\top\hat B)^{-1}\hat B^\top$ and $\|\hat B(\hat B^\top\hat B)^{-1}\|_2=1/\sigma_{\min}(\hat B)$. For the feature map, Eq.~\eqref{Equ:jacobian} gives the Jacobian $Fs\,(I-\tfrac12s^2\pi h^\top)$ with $s=(\pi^\top h+\zeta)^{-1/2}\le\zeta^{-1/2}$, because $h\ge0$ entrywise and $\pi\ge0$, and $\|\pi h^\top\|_2\le\|h\|_2$ on $\Delta_d$, so its operator norm is at most $L_\phi$ on the convex set $\Delta_d$ and the mean value inequality applies. Replacing $(\Pi^\star,C^\star)$ by $(\Pi^\star S,S^{-1}C^\star S^{-\top})$ maps $\pi^\star_v\mapsto S^\top\pi^\star_v$ and $B^\star\mapsto B^\star S^{-\top}$, leaves $\rho_v$ and $P^\star$ unchanged, and the argument applies verbatim.
\end{proof}
The first term is specific to the node: its numerator $\hat B^\top\rho_v=\sum_{u\in\mathcal R}\hat g_u\rho_{uv}$ sums the residual of the reference fit at the revealed entries of $v$ against the landmark directions, and its denominator $\sigma_{\min}(\hat B)^2=\lambda_{\min}\bigl(\sum_{u\in\mathcal R}\hat g_u\hat g_u^\top\bigr)$ is nondecreasing when directions are added with the existing directions held fixed. In the class $\rho_v=0$, and after identification $\hat B=B^\star S^{-\top}$ for a gauge $S$ by Proposition~\ref{prop:coverage}, so both terms vanish. The second term, the error of the landmark directions, is shared by all nodes, as is the error of the global quantities $(R,V,h)$ that the proposition holds fixed. Under pair queries the same argument applies with $\hat B$ replaced by the $|R_v|\times d$ matrix of directions against which $v$ was probed. With exactly $d$ probes that matrix is square, $P_{\hat B}=I$, and the bound reads $\|\rho_v\|_2/\sigma_{\min}(\hat B)$, so the residual passes through without projection, whereas every additional row projects it onto a $d$-dimensional subspace of a growing space (Appendix~\ref{app:exp-patterns}). Section~\ref{sec:exp-q2} measures the total gap, shared errors included.

\paragraph{The oracle approximant and its bias.}
Nothing about $A$ is assumed by the method. The price of finite resolution is measured by the deterministic bias functional
\begin{equation}
\label{Equ:bias}
\varepsilon_{d}(A):=\min_{X\in\mathcal C_{d}}\|(A-X)_{\mathrm{off}}\|_F,
\qquad
P^\star\in\arg\min_{X\in\mathcal C_{d}}\|(A-X)_{\mathrm{off}}\|_F,
\end{equation}
where $(\cdot)_{\mathrm{off}}$ zeroes the diagonal, since self-pairs are never queried. The minimum is attained (Appendix~\ref{app:derivations}(B)), and a fixed minimizer $P^\star$ is the reference surrogate of Section~\ref{sec:theory}. Proposition~\ref{prop:node-error} shows how its residuals enter the features. The bias is non-increasing in $d$, vanishes when $A_{\mathrm{off}}$ lies in the class, and decays as $d^{-\alpha/\ell}$ when $A_{ij}=f(x_i,x_j)$ for $\alpha$-H\"older $f$ and latent positions in $[0,1]^\ell$ (Lemma~\ref{lem:bias}, Appendix~\ref{app:method}). For a sparse binary network the relevant comparison is between $L_\zeta(\tilde A)$ and $L_\zeta(A)$ in operator norm, which degree regularization with $\zeta$ of the order of the mean degree keeps controlled~\citep{chaudhuri2012spectral,qin2013regularized,le2017concentration}.

\section{Method Details}
\label{app:method}
This appendix gives Algorithm~\ref{Alg:global_bo}, the round summarized in Section~\ref{sec:method}, together with the coefficient update, local feature model, exact spectral solve, and query implementation used there.

\begin{algorithm}[t]
\caption{Global BO on a Partially Observed Graph (one campaign)}
\label{Alg:global_bo}
\begin{algorithmic}[1]
\STATE \textbf{Input:} $\mathcal V$, $T$, $N_0$, resolution $d$, per-round budget of $Q_R$ rows or $Q_E$ pairs, pin share $\rho_{\mathrm{pin}}$
\STATE \textbf{Initialize:} evaluate $y$ on $N_0$ random nodes and reveal their rows or their pair block (the landmarks)
\FOR{$t=0,\dots,T-1$}
  \STATE \textbf{Pin:} reveal the rows, or $d$ landmark probes, of unpinned evaluated nodes within the share $\rho_{\mathrm{pin}}$
  \STATE \textbf{Rows:} spend the remaining rows uniformly at random over unrevealed nodes (Algorithm~\ref{Alg:structure})
  \STATE \textbf{Fit:} solve~\eqref{Opt:fit} and compute the spectrum, features $\phi_v$, and kernel~\eqref{Equ:kernel2} exactly (Lemma~\ref{lem:exact-solve})
  \STATE \textbf{Posterior:} $\Sigma_v,\Sigma_v^\phi$ by~\eqref{Equ:info}, $(w_t,\Xi_t)$ by Lemma~\ref{lem:reduction} with~\eqref{Equ:eiv} at unpinned evaluated nodes, $U_t$ by~\eqref{Equ:ucb-struct}
  \STATE \textbf{Pairs:} spend the remaining pairs one at a time by $\mathrm{KG}^\ast_t$~\eqref{Equ:kg}, refreshing the queried node (Algorithm~\ref{Alg:pair})
  \STATE \textbf{Evaluate:} $v^{(t+1)}=\arg\max_{v\notin\mathcal V^{(t)}}U_t(v)$, observe $y(v^{(t+1)})$, and add it to $\mathcal V^{(t+1)}$
\ENDFOR
\STATE \textbf{return} $\hat v=\arg\max_{v\in\mathcal V^{(T)}}y(v)$
\end{algorithmic}
\end{algorithm}

\paragraph{Coefficient posterior.}
The structured-plus-bulk kernel belongs to
\begin{equation}
\label{Equ:family}
\mathcal K(\tilde U)=\{\sigma_b^2I+\tilde U\operatorname{diag}(a)\tilde U^\top:\sigma_b^2>0,\ a\ge0\}.
\end{equation}
\begin{lemma}[Reduction to a linear model in node features]
\label{lem:reduction}
Let $\tilde K=\sigma_b^2I+\tilde U D\tilde U^\top\in\mathcal K(\tilde U)$ with $D\succ0$ diagonal, $\tilde U$ collecting the $d_+$ modes with $a_i>0$, let $\phi_v$ be the transposed $v$-th row of $\tilde U$, $S=\mathcal V^{(t)}$, and $G_S=\tilde U_S^\top\tilde U_S$. For every $v\notin S$,
\begin{equation}
\label{Equ:reduction}
\begin{aligned}
\mu_t(v)&=\phi_v^\top w_t, & w_t&=\bigl(\sigma_b^2D^{-1}+G_S\bigr)^{-1}\tilde U_S^\top\mathbf y_S,\\
s_t^2(v)&=\sigma_b^2+\phi_v^\top\Xi_t\phi_v, & \Xi_t&=\sigma_b^2\bigl(\sigma_b^2D^{-1}+G_S\bigr)^{-1},
\end{aligned}
\end{equation}
the posterior of the Bayesian linear model $y(v)=\langle\phi_v,\theta\rangle+\varepsilon_v$, $\theta\sim\mathcal N(0,D)$, $\varepsilon_v\sim\mathcal N(0,\sigma_b^2)$. Exact inference over all $n$ nodes costs $\mathcal O(|S|d^2+nd^2+d^3)$ per round (Appendix~\ref{app:derivations}(D)).
\end{lemma}

\paragraph{Feature map and local covariance.}
Use the factors of Lemma~\ref{lem:exact-solve}, and let $V_+,\tilde\Gamma_+$ retain the positive eigenvalues of the normalized adjacency. With fitted global quantities fixed,
\begin{equation}
\label{Equ:feature}
\phi_v=\frac{F\pi_v}{\sqrt{\pi_v^\top h+\zeta}},\qquad
F=\tilde\Gamma_+^{-1}V_+^\top RC,\qquad h=C\Pi^\top\mathbf1.
\end{equation}
Indeed $\tilde U_+=MCR^\top V_+\tilde\Gamma_+^{-1}$. If $R$ is invertible, $F=V_+^\top R^{-\top}$, recovering the full-rank form. The first expression also applies when the memberships are rank deficient. Its local Jacobian is
\begin{equation}
\label{Equ:jacobian}
J_v=F s_v\bigl(I-\tfrac12s_v^2\pi_vh^\top\bigr),\qquad
s_v=(\pi_v^\top h+\zeta)^{-1/2}.
\end{equation}
For $G_v=[g_u^\top]_{u\in R_v}$, with $R_v$ the measured landmarks of $v$, Gaussian conditioning gives
\begin{equation}
\label{Equ:info}
\Sigma_v=\Sigma_0-\Sigma_0G_v^\top(G_v\Sigma_0G_v^\top)^+G_v\Sigma_0,
\qquad \Sigma_v^\phi=J_v\Sigma_vJ_v^\top.
\end{equation}
Here $\Sigma_0$ is the empirical covariance of the landmark memberships plus a positive ridge. These covariances describe unmeasured directions conditional on the fit. In the exact, consistent measurement model, the conditional mean can be projected onto $\{\pi\in\Delta_d:G_v\pi=A_{R_v,v}\}$ when this set is nonempty. For general observations the constrained residual fit~\eqref{Opt:fit} supplies the membership estimate.

\paragraph{Exact spectral solve.}
\begin{lemma}[Exact spectral solve]
\label{lem:exact-solve}
Let $\tilde A=\Pi C\Pi^\top\in\mathcal C_{d}$ with $d\le n$ and
$\zeta>0$. Let $M=(\tilde D+\zeta I)^{-1/2}\Pi$ with thin QR factorization
$M=QR$, and let $RCR^\top=V\tilde\Gamma V^\top$ be a $d\times d$
eigendecomposition. Then $\tilde U=QV$ collects orthonormal eigenvectors
of $\tilde L_\zeta$ with eigenvalues $\tilde\lambda_i=1-\tilde\gamma_i$,
and all remaining eigenvalues of $\tilde L_\zeta$ equal $1$. The total
cost is $\mathcal O(nd^2+d^3)$.
\end{lemma}
\begin{proof}
$\tilde L_\zeta=I-(\tilde D+\zeta I)^{-1/2}\Pi C\Pi^\top(\tilde D+\zeta I)^{-1/2}=I-MCM^\top=I-QRCR^\top Q^\top=I-QV\tilde\Gamma V^\top Q^\top$. The columns of $\tilde U=QV$ are orthonormal, so $\tilde L_\zeta\tilde U=\tilde U(I-\tilde\Gamma)$, which gives the $d$ eigenpairs. For $x$ orthogonal to $\operatorname{col}(Q)$ we have $M^\top x=R^\top Q^\top x=0$, so $\tilde L_\zeta x=x$ and every remaining eigenvalue equals $1$. The thin QR costs $\mathcal O(nd^2)$, the $d\times d$ eigendecomposition $\mathcal O(d^3)$, and the product $QV$ another $\mathcal O(nd^2)$.
\end{proof}

\paragraph{Predictive variance under feature uncertainty.}
Equation~\eqref{Equ:three-term} follows from a moment calculation conditional on the current data. Under the working independence of $\phi_v$ and $\theta$,
\[
\operatorname{Var}(\phi_v^\top\theta)
=\mathbb E[\phi_v^\top\Xi_t\phi_v]
+\operatorname{Var}(\phi_v^\top w_t).
\]
The first term is $\hat\phi_v^\top\Xi_t\hat\phi_v+\operatorname{tr}(\Sigma_v^\phi\Xi_t)$ and the second is $w_t^\top\Sigma_v^\phi w_t$. Adding the variance $\sigma_b^2$ of an independent residual gives the three terms in the main text. Conditional independence and the local propagation $\Sigma_v^\phi=J_v\Sigma_vJ_v^\top$ are working approximations that hold the fitted global quantities fixed. Node selection uses the plug-in score of Eq.~\eqref{Equ:ucb-struct}, and this variance prices structural queries.
For an evaluated node whose feature is still uncertain, we instead use the moment-based errors-in-variables adjustment
\begin{equation}
\label{Equ:eiv}
\sigma_b^2(v)=\sigma_b^2+w_t^\top\Sigma^\phi_vw_t+\operatorname{tr}\bigl(\Sigma^\phi_v\Xi_t\bigr)
\end{equation}
in place of $\sigma_b^2$ when estimating the shared coefficients. This is an approximate treatment of feature error, whose effect on the coefficient estimate is described in Appendix~\ref{app:derivations}(I).

\paragraph{Heteroscedastic form of Lemma~\ref{lem:reduction}.}
With node-wise variances $N_S=\operatorname{diag}(\sigma_b^2(v))_{v\in S}$ from Eq.~\eqref{Equ:eiv}, the GP state is $\Xi_t=(D^{-1}+\tilde U_S^\top N_S^{-1}\tilde U_S)^{-1}$ and $w_t=\Xi_t\tilde U_S^\top N_S^{-1}\mathbf y_S$, which reduces to Eq.~\eqref{Equ:reduction} when $N_S=\sigma_b^2I$.

\paragraph{Rank-one update of the design covariance.}
One more measurement of $v$ along a direction $g$ updates $\Sigma_v$ by
\begin{equation}
\label{Equ:rankone}
\Sigma_v'=\Sigma_v-\frac{\Sigma_v\,gg^\top\Sigma_v}{g^\top\Sigma_v\,g},
\end{equation}
which depends on $g$ only through its component outside the row space of $G_v$ and is void along an already measured direction.

\paragraph{Spectral containment.}
\begin{lemma}[Spectral containment]
\label{lem:containment}
If $\tilde A\ge0$ entrywise and $\zeta>0$, then
$\operatorname{spec}(\tilde L_\zeta)\subset[\zeta/(\widetilde{\deg}_{\max}+\zeta),\,2-\zeta/(\widetilde{\deg}_{\max}+\zeta)]$. In particular $\tilde
L_\zeta\succ0$. For the resolvent $r(\lambda)=(\beta+\lambda)^{-1}$ with $\beta>0$, $r(\tilde L_\zeta)\succeq(\beta+2)^{-1}I$. The structured-plus-bulk kernel of Eq.~\eqref{Equ:kernel2} satisfies $\tilde K\succeq\sigma_b^2I$.
\end{lemma}
The lower end is the quadratic-form argument of Appendix~\ref{app:derivations}(A) applied to $\tilde A$, whose diagonal is included in $\widetilde{\deg}$. For the upper end, with $\breve x=(\tilde D+\zeta I)^{-1/2}x$, $x^\top(2I-\tilde L_\zeta)x=\|x\|_2^2+\breve x^\top\tilde A\breve x=\tfrac12\sum_{i,j}\tilde A_{ij}(\breve x_i+\breve x_j)^2+\zeta\|\breve x\|_2^2\ge\zeta(\widetilde{\deg}_{\max}+\zeta)^{-1}\|x\|_2^2$, since $\|x\|_2^2=\breve x^\top(\tilde D+\zeta I)\breve x$. The resolvent bound follows from $\tilde L_\zeta\preceq2I$, and $\tilde K\succeq\sigma_b^2I$ because the structured term is positive semidefinite.

\paragraph{One-scale kernels and the linear model.}
For a response positive on the spectrum, the exact one-scale kernel has the identity
\begin{equation}
\label{Equ:kernel}
K_{\mathrm{one}}=r(\tilde L_\zeta)=r(1)I+\tilde U D_r\tilde U^\top,
\qquad D_r=\operatorname{diag}(r(\tilde\lambda_i)-r(1)).
\end{equation}
It uses the same surrogate and spectral solve, and supports standard GP conditioning. When $D_r\succeq0$, removing zero corrections gives Lemma~\ref{lem:reduction} with bulk variance $r(1)$. If $D_r$ has negative entries, the kernel remains positive definite but these entries cannot be variances of independent coefficients. Its positive decomposition instead uses the correlated residual $r(1)(I-\tilde U\tilde U^\top)$ and structured term $\tilde U\operatorname{diag}(r(\tilde\lambda_i))\tilde U^\top$.

Clipping negative corrections gives $K_{\mathrm{clip}}=r(1)I+\tilde U_+D_+\tilde U_+^\top$, the tied-scale case of Eq.~\eqref{Equ:kernel2}. Our chosen kernel additionally separates the scales. This allows the strength of the structured component to vary independently of the bulk, whose relative average correction in the one-scale kernel is $\mathcal O(d/n)$ when spectral contrasts are bounded (Appendix~\ref{app:derivations}(E)). Clipping and scale separation are modeling choices, while Eq.~\eqref{Equ:kernel} is an identity. Appendix~\ref{sec:ablation-scales} compares the chosen kernel with $K_{\mathrm{one}}$.

\paragraph{Bias under smooth latent interaction.}
The following lemma gives an interpretable sufficient condition for a small bias $\varepsilon_d(A)$, namely smooth latent interaction, with a fully deterministic proof in Appendix~\ref{app:derivations}(C). It is the noise-free weighted analogue of the eigenvalue-tail estimates behind spectral graphon estimation~\citep{xu2018rates,chatterjee2015usvt}.

\begin{lemma}[Bias under smooth latent interaction]
\label{lem:bias}
Suppose $A_{ij}=f(x_i,x_j)$ for all $i\ne j$, with latent positions
$x_i\in[0,1]^{\ell}$ and $f:[0,1]^{\ell}\times[0,1]^{\ell}\to[0,1]$ symmetric and
$\alpha$-H\"older with constant $L$ in each coordinate, $\alpha\in(0,2]$
(for $\alpha\in(1,2]$: differentiable with $(\alpha{-}1)$-H\"older partial
derivatives). Then for every $d\ge 2^{\ell}$ of the form $d=(k{+}1)^\ell$,
\begin{equation*}
\varepsilon_{d}(A)\;\le\; 2^{\alpha+1} \ell\, L\, n\, d^{-\alpha/\ell}.
\end{equation*}
The witness is explicit and lies in the class. The row $\pi_i$ collects
the tensor hat-function (degree-one B-spline) basis evaluated at $x_i$,
which is nonnegative and sums to one, and $C_{ab}=f(t_a,t_b)\in[0,1]$ are
function values at the knots. Hard-membership blocks, the step-graphon
subclass, already achieve the rate for $\alpha\le1$, and the
soft-membership construction extends it to $\alpha\le2$. For general
$d$, monotonicity of $\varepsilon_{d}$ in $d$ gives the same rate up
to the constant factor $2^{\alpha}$ by rounding down to the nearest
$\ell$-th power.
\end{lemma}

The hard-membership subclass of $\mathcal C_{d}$ consists of $d$-block step matrices, which connects the class to stochastic block models and step graphons~\citep{airoldi2008mixed,lovasz2012large} as descriptions of the approximant rather than as assumptions on $A$. An indefinite core $C$ represents disassortative (heterophilous) structure.

\begin{algorithm}[t]
\caption{Surrogate Reconstruction and Kernel Construction}
\label{Alg:kernel_learning}
\textbf{Input:} observed entries $\Omega$ with landmarks $\mathcal R$, resolution $d$, regularizers
$(\zeta,\beta)$ and scales $(\sigma_b^2,\sigma_s^2)$, previous fit $(\Pi,C)$ if available, refit flag\\
\textbf{Output:} kernel $\tilde k(\cdot,\cdot)$ in factored form and feature covariances $\{\Sigma^\phi_v\}$
\begin{algorithmic}[1]
\STATE \textbf{Structured initialization} (first call, or refit flag set): $(\Pi_{\mathcal R},C)\gets$ rank-$d$ factorization of the observed landmark entries by truncated SVD or successive projection, and each remaining $\pi_v\gets$ projected least squares on its observed entries
\hfill$\triangleright$ $\mathcal O(|\Omega|d)$
\REPEAT
  \STATE one alternating least-squares sweep on Problem~\eqref{Opt:fit}: each membership by projected least squares over $\Delta_d$ given the others and $C$, then $C$ by box-constrained least squares given $\Pi$ (Appendix~\ref{app:coldstart})
  \hfill$\triangleright$ $\mathcal O(|\Omega|d+nd^2)$ per sweep under rows, $\mathcal O(|\Omega|d^2+nd^2)$ under pairs
\UNTIL{the sweep budget is spent or the residual decrease falls below a tolerance}
\STATE If the refit flag is set, run the sweeps from the previous fit as well and keep the solution with the smaller residual
\STATE $\widetilde{\deg}\gets\Pi(C(\Pi^\top\mathbf 1))$
\hfill$\triangleright$ $\mathcal O(nd^2+|\Omega|d)$
\STATE $M\gets(\tilde D+\zeta I)^{-1/2}\Pi$, thin QR $M=QR$, eigendecomposition $RCR^\top=V\tilde\Gamma V^\top$, $\tilde U\gets QV$, $\tilde\Lambda\gets I-\tilde\Gamma$; keep the $d_+$ directions with $\tilde\lambda_i<1$
\hfill$\triangleright$ Lemma~\ref{lem:exact-solve}, $\mathcal O(nd^2+d^3)$
\STATE $\Sigma_v\gets$ Eq.~\eqref{Equ:info} with $g_u=C\pi_u$, and $\Sigma^\phi_v\gets J_v\Sigma_vJ_v^\top$
\hfill$\triangleright$ membership covariance shared under rows (Proposition~\ref{prop:shared})
\STATE \textbf{return} $\tilde k$ via Eq.~\eqref{Equ:kernel2} as
$(\sigma_b^2,\tilde U_+,\sigma_s^2D_+)$, and $\{\Sigma^\phi_v\}$
\hfill$\triangleright$ $\mathcal O(nd)$ storage
\end{algorithmic}
\end{algorithm}

\begin{table}[t]
\centering
\caption{Per-round complexity with $m=|\mathcal V^{(t)}|$ queried nodes and $r=|\mathcal R^{(t)}|$ landmarks.
By Lemma~\ref{lem:reduction} the GP state is $(G_S,\tilde U_S^\top\mathbf y_S)$, updated in $\mathcal O(d^2)$ per new evaluation while features are fixed. After one $d\times d$ solve, posterior means over all $n$ nodes cost $\mathcal O(nd)$ and exact variances $\mathcal O(nd^2)$, independent of $m$.}
\label{tab:complexity}
\small
\begin{tabular}{lc}
\toprule
storage (factors, spectrum, GP state) & $\mathcal O(nd+|\Omega|+d^2)$\\
surrogate update (per ALS sweep, rows / pairs) & $\mathcal O(|\Omega|d+nd^2)$ / $\mathcal O(|\Omega|d^2+nd^2)$\\
spectral refresh (exact) & $\mathcal O(nd^2+d^3)$\\
GP state update (Lemma~\ref{lem:reduction}) & $\mathcal O(md^2+d^3)$\\
acquisition: means (given $w_t$) & $\mathcal O(nd)$\\
acquisition: exact variances & $\mathcal O(nd^2)$\\
local covariance propagation under rows (Eq.~\eqref{Equ:info}) & $\mathcal O(rd^2+nd^2)$\\
row acquisition (Algorithm~\ref{Alg:structure}) & $\mathcal O(nrd)$\\
pair acquisition, per query (Algorithm~\ref{Alg:pair}) & $\mathcal O(rd^3+n)$\\
\bottomrule
\end{tabular}
\end{table}

\paragraph{Row rule.}
\begin{proposition}[Shared membership uncertainty under row queries]
\label{prop:shared}
With common reference covariance $\Sigma_0$ and fitted directions held fixed, let $G_{\mathcal R}=[g_u^\top]_{u\in\mathcal R}$ for the revealed rows. Every unrevealed node has the same covariance $\Sigma_{\mathcal R}$ given by Eq.~\eqref{Equ:info}. If $\operatorname{rank}G_{\mathcal R}=d$, this covariance is zero.
\end{proposition}
The proof is in Appendix~\ref{app:derivations}(F). Feature and score uncertainties also depend on $J_v$ and $c_v$. Algorithm~\ref{Alg:structure} first spends its pinning allowance on evaluated nodes, then draws uniformly from the remaining rows. The uniform component provides broad coverage, while the pinning component improves the features used to estimate the shared coefficient.

\begin{algorithm}[t]
\caption{Row Acquisition (one round)}
\label{Alg:structure}
\textbf{Input:} revealed rows $\mathcal R$, evaluated nodes $\mathcal V^{(t)}$, number of rows $q_t$ available in this round, pin share $\rho_{\mathrm{pin}}$, fit $(\Pi,C)$, resolution $d$\\
\textbf{Output:} updated $\mathcal R$ and $\Omega$
\begin{algorithmic}[1]
\STATE $q\gets0$
\WHILE{$q<q_t$, the pin share is not exhausted, and some evaluated node has an unrevealed row}
  \STATE reveal the row of the earliest such node and set $q\gets q+1$
  \hfill$\triangleright$ accurate features at evaluated nodes
\ENDWHILE
\WHILE{$q<q_t$}
  \STATE draw $w$ uniformly at random from $\mathcal V\setminus\mathcal R$
  \hfill$\triangleright$ uniform rows, Proposition~\ref{prop:shared}
  \STATE reveal the row of $w$, set $\mathcal R\gets\mathcal R\cup\{w\}$ and $q\gets q+1$, and update the fit by warm-started sweeps of Algorithm~\ref{Alg:kernel_learning}
\ENDWHILE
\STATE \textbf{return} $\mathcal R$ and the corresponding $\Omega$
\end{algorithmic}
\end{algorithm}

\section{Derivations for the Methodology}
\label{app:derivations}

\paragraph{(A) Dirichlet-energy identity.}
Let $A\ge0$ be symmetric with $A_{ii}=0$, $W=(D_A+\zeta I)^{-1/2}$,
$\breve y=Wy$. Then $\|y\|^2_{\mathcal H(K(A))}=y^\top K(A)^{-1}y
=y^\top(\beta I+L_\zeta(A))y=\beta\|y\|_2^2+y^\top L_\zeta(A)y$. Since
$y^\top L_\zeta(A)y=\|y\|_2^2-\breve y^\top A\breve y$ and
$\|y\|_2^2=\breve y^\top(D_A+\zeta I)\breve y
=\breve y^\top D_A\breve y+\zeta\|\breve y\|_2^2$, we get
$y^\top L_\zeta(A)y=\breve y^\top(D_A-A)\breve y+\zeta\|\breve y\|_2^2$.
With $\deg_i=\sum_j A_{ij}$,
$\breve y^\top D_A\breve y=\sum_{ij}A_{ij}\breve y_i^2$, so by symmetry
$\breve y^\top(D_A-A)\breve y
=\tfrac12\sum_{ij}A_{ij}(\breve y_i-\breve y_j)^2
=\tfrac12\sum_{i\ne j}A_{ij}(\breve y_i-\breve y_j)^2$
(diagonal terms vanish, so the identity also holds when $A$ has a nonnegative diagonal that $D_A$ includes, as for the surrogate $\tilde A$). The same computation applied to
$x^\top L_\zeta(A)x$ shows $L_\zeta(A)\succeq
\zeta/(\deg_{\max}(A)+\zeta)\,I\succ0$. {Applied to the surrogate, the identity shows that a signal $x$ with $\breve x^\top\tilde A\breve x=0$, uncorrelated along the edges of the surrogate, has Rayleigh quotient $x^\top\tilde L_\zeta x/\|x\|^2=1$, which is the reading of the eigenvalue $1$ used in Section~\ref{Subsec:global BO}.}

\paragraph{(B) Attainment of the bias minimum \eqref{Equ:bias}.}
$(\Delta_{d})^n\times\{C=C^\top\in[0,1]^{d\times d}\}$ is compact,
$(\Pi,C)\mapsto\Pi C\Pi^\top$ is continuous, so $\mathcal C_{d}$ is
compact. $X\mapsto\|(A-X)_{\mathrm{off}}\|_F$ is continuous, so the minimum is attained by the Weierstrass theorem.

\paragraph{(C) Proof of Lemma~\ref{lem:bias}.}
\emph{Univariate pieces.} Let $g:[0,1]\to[0,1]$ and let $I_h$ denote
piecewise-linear (hat-function) interpolation at knots $t_a=a/k$,
$h=1/k$. (i) If $|g(s)-g(t)|\le L|s-t|^\alpha$, $\alpha\in(0,1]$: for
$x\in[t_a,t_{a+1}]$, $I_hg(x)=\lambda g(t_a)+(1-\lambda)g(t_{a+1})$ is a
convex combination, so $|g(x)-I_hg(x)|\le\lambda|g(x)-g(t_a)|
+(1-\lambda)|g(x)-g(t_{a+1})|\le Lh^\alpha$. (ii) If $g$ is differentiable
with $|g'(s)-g'(t)|\le L|s-t|^{\alpha-1}$, $\alpha\in(1,2]$: on $[a,b]$
with $h=b-a$, the mean value theorem gives
$g(b)-g(a)=hg'(\xi)$ for some $\xi\in(a,b)$, hence
$g(x)-I_hg(x)=\int_a^x\bigl(g'(t)-g'(\xi)\bigr)\,dt$ and
$|g(x)-I_hg(x)|\le(x-a)\,L\,h^{\alpha-1}\le Lh^\alpha$.

\emph{Tensorization.} View $f$ as a function of $2\ell$ scalar coordinates
and let $I^{\otimes}=I_1\circ\cdots\circ I_{2\ell}$ interpolate each
coordinate on the same knot grid. Each coordinate operator is positive with
partition of unity, hence sup-norm nonexpansive, so telescoping
$f-I^{\otimes}f=\sum_{j=1}^{2\ell}(I_1\cdots I_{j-1})(f-I_jf)$ yields
$\|f-I^{\otimes}f\|_\infty\le\sum_j\|f-I_jf\|_\infty\le 2\ell\,Lh^\alpha$.

\emph{Class membership.} $I^{\otimes}f(x,y)=\sum_{a,b}f(t_a,t_b)\,
\varphi_a(x)\varphi_b(y)$, where $\{\varphi_a\}$ is the tensor hat basis
over $[0,1]^\ell$ with $(k{+}1)^\ell$ elements: $\varphi_a\ge0$,
$\sum_a\varphi_a\equiv1$. Setting $\pi_i=(\varphi_a(x_i))_a\in\Delta_{d}$
and $C_{ab}=f(t_a,t_b)\in[0,1]$ (symmetric since $f$ is) exhibits
$X=\Pi C\Pi^\top\in\mathcal C_{d}$ with $d=(k{+}1)^\ell$ and
$X_{ij}=I^{\otimes}f(x_i,x_j)$.

\emph{Constants.} For $d\ge2^\ell$, $k=d^{1/\ell}-1\ge d^{1/\ell}/2$, so
$h\le 2d^{-1/\ell}$ and the entrywise error is at most
$2\ell L(2d^{-1/\ell})^\alpha\le 2^{\alpha+1}\ell L\,d^{-\alpha/\ell}$. Summing over
at most $n^2$ off-diagonal entries gives
$\varepsilon_{d}(A)\le 2^{\alpha+1}\ell Ln\,d^{-\alpha/\ell}$.
For $\alpha\le1$, replacing hats by indicators of $d=k^\ell$ congruent
cells (hard memberships, $C_{ab}=f$ at cell centers) gives
$\varepsilon_{d}\le 2^{1-\alpha}\ell Ln\,d^{-\alpha/\ell}$, proving the
step-graphon claim. \hfill$\square$

\paragraph{(D) Proof of Lemma~\ref{lem:reduction}.}
Let $\tilde K=\sigma_b^2I+\tilde UD\tilde U^\top$ and $v\notin S$, so that $\tilde K_{vS}=\phi_v^\top D\tilde U_S^\top$, $\tilde K_{vv}=\sigma_b^2+\phi_v^\top D\phi_v$, and $\tilde K_{SS}=\sigma_b^2I_m+\tilde U_SD\tilde U_S^\top$. The push-through identity $B(\sigma_b^2I_m+AB)^{-1}=(\sigma_b^2I_d+BA)^{-1}B$ with $B=D\tilde U_S^\top$ and $A=\tilde U_S$ gives $\tilde K_{vS}\tilde K_{SS}^{-1}=\phi_v^\top(\sigma_b^2I_d+DG_S)^{-1}D\tilde U_S^\top=\phi_v^\top(\sigma_b^2D^{-1}+G_S)^{-1}\tilde U_S^\top$, hence $\mu_t(v)=\phi_v^\top w_t$. By the Woodbury identity, $D-D\tilde U_S^\top(\sigma_b^2I_m+\tilde U_SD\tilde U_S^\top)^{-1}\tilde U_SD=(D^{-1}+G_S/\sigma_b^2)^{-1}=\sigma_b^2(\sigma_b^2D^{-1}+G_S)^{-1}=\Xi_t$, hence $s_t^2(v)=\sigma_b^2+\phi_v^\top\Xi_t\phi_v$. For $D\succ0$ the same expressions are the posterior of $\theta$ in $y=\langle\phi,\theta\rangle+\varepsilon$ with prior $\mathcal N(0,D)$ and per-node bulk variance $\sigma_b^2$. For fixed features, $G_S$ and $\tilde U_S^\top\mathbf y_S$ are updated in $\mathcal O(d^2)$ per new evaluation. One $d\times d$ solve then gives $w_t$ and $\Xi_t$, and each node costs $\mathcal O(d)$ for the mean and $\mathcal O(d^2)$ for the variance. \hfill$\square$

\paragraph{(E) Contrast bound.}
\begin{lemma}[Contrast bound]
\label{lem:contrast-bound}
For $\tilde K=r(\tilde L_\zeta)$ with $r>0$ nonincreasing, $\frac1n\sum_v(\tilde K_{vv}-r(1))/r(1)\le\frac dn\,(r(\tilde\lambda_{\min})-r(1))/r(1)$. For the resolvent the right-hand side is at most $\frac dn(1-\tilde\lambda_{\min})/\tilde\lambda_{\min}$ for every $\beta>0$, and for the heat filter $e^{-\tau\lambda}$ it equals $\frac dn(e^{\tau(1-\tilde\lambda_{\min})}-1)$, which is unbounded in $\tau$.
\end{lemma}
Proof: $\tilde K_{vv}=r(1)+\sum_{i\le d}(r(\tilde\lambda_i)-r(1))\tilde U_{vi}^2$ and $\sum_v\tilde U_{vi}^2=1$. For the resolvent, $(r(\tilde\lambda)-r(1))/r(1)=(1-\tilde\lambda)/(\beta+\tilde\lambda)$. \hfill$\square$

Example: four disconnected cliques of $250$ nodes, $\zeta=1$, true graph. Within a clique $\tilde L_\zeta$ has the eigenvalue $1-249/250=0.004$ on the clique indicator and $1+1/250=1.004$ on its complement. With $r(\lambda)=(\beta+\lambda)^{-1}$ at $\beta=0.2$, $r(0.004)=4.90$ and $r(1.004)=0.831$, so the within-clique kernel correlation is $(4.90-0.831)/(4.90+249\cdot0.831)=0.019$. As $\beta\to0$ it tends to $(250-0.996)/(250+249\cdot0.996)=0.50$, and with the heat filter $e^{-10\lambda}$ it is $0.989$. The same $d/n$ dilution holds for the true graph and for the surrogate, since it comes from having few structured modes and many bulk modes and not from the low-rank class.

\paragraph{(F) Shared covariance and the weighted query gain.}
For Proposition~\ref{prop:shared}, Eq.~\eqref{Equ:info} depends on a node only through $G_v$, which is $G_{\mathcal R}$ for every unrevealed node. If $G_{\mathcal R}$ has full column rank, exact conditioning leaves zero covariance. A further direction $g$ induces Eq.~\eqref{Equ:rankone}, which depends only on its component outside the row space of $G_{\mathcal R}$. These statements concern membership covariance conditional on the fit, while its propagation to scores also uses each node's sensitivity. \hfill$\square$

The gain $g\mapsto(c^\top\Sigma g)^2/(g^\top\Sigma g)$ of Appendix~\ref{app:pair} is not submodular as a set function of the probes. With $d=2$, $\Sigma=I$, $c=e_1$, $g_1=e_2$, $g_2=e_1+e_2$, the marginal gain of $g_1$ is $0$ alone and $1/2$ after $g_2$, because after $g_2$ one has $\Sigma'=I-g_2g_2^\top/2$, $c^\top\Sigma'g_1=-1/2$, and $g_1^\top\Sigma'g_1=1/2$.

\paragraph{(G) Proof of Proposition~\ref{prop:coverage}.}
Two facts are used repeatedly. \emph{Fact 1 (open image).} Let $\mathcal S$ be a node set, let the representative be interior with $\operatorname{rank}\Pi_{\mathcal S}=\min(|\mathcal S|,d)$, and let $\mathcal M_{\mathcal S}$ be the manifold of symmetric $|\mathcal S|\times|\mathcal S|$ matrices of rank $\min(|\mathcal S|,d)$, of dimension $|\mathcal S|d-\binom d2$ if $|\mathcal S|\ge d$ and $\binom{|\mathcal S|+1}2$ otherwise. The map $(\Pi_{\mathcal S},C)\mapsto\Pi_{\mathcal S}C\Pi_{\mathcal S}^\top$ has differential $(\dot\Pi,\dot C)\mapsto\dot\Pi C\Pi_{\mathcal S}^\top+\Pi_{\mathcal S}\dot C\Pi_{\mathcal S}^\top+\Pi_{\mathcal S}C\dot\Pi^\top$, which is onto the tangent space of $\mathcal M_{\mathcal S}$ at $\Pi_{\mathcal S}C\Pi_{\mathcal S}^\top$, so by the submersion theorem the image of a neighborhood of the representative, which stays inside the class because the representative is interior, contains an open subset of $\mathcal M_{\mathcal S}$. A set $\Omega$ of entries of the block defines a continuous map from that open subset, homeomorphic to an open subset of $\mathbb R^{\dim\mathcal M_{\mathcal S}}$, into $\mathbb R^{|\Omega|}$, and if $|\Omega|<\dim\mathcal M_{\mathcal S}$ this map is not injective, by invariance of domain, so two members of the class share the observations. \emph{Fact 2 (skeleton).} If $X$ is symmetric of rank at most $d$, $X_{\mathcal R,\cdot}=M_{\mathcal R,\cdot}$, and $\operatorname{rank}M_{\mathcal R\mathcal R}=d$, then $X=M$. Indeed $X$ contains the rank-$d$ block $X_{\mathcal R\mathcal R}=M_{\mathcal R\mathcal R}$, so $X$ has rank exactly $d$ and its row space is that of $X_{\mathcal R,\cdot}$, hence $X=WX_{\mathcal R,\cdot}$ for some $W$ and $X_{\cdot,\mathcal R}=WX_{\mathcal R\mathcal R}$. Since $X_{\mathcal R\mathcal R}X_{\mathcal R\mathcal R}^{+}$ is the orthogonal projector onto $\operatorname{col}(X_{\mathcal R\mathcal R})=\operatorname{col}(X_{\mathcal R,\cdot})$, $X=X_{\cdot,\mathcal R}X_{\mathcal R\mathcal R}^{+}X_{\mathcal R,\cdot}$, the same identity holds for $M$, and the three factors agree because $X_{\cdot,\mathcal R}=X_{\mathcal R,\cdot}^\top=M_{\cdot,\mathcal R}$.

\emph{(i).} Fact 1 with $\mathcal S=\mathcal V$.

\emph{(ii), sufficiency.} Let $B=\Pi_{\mathcal R}C\in\mathbb R^{r\times d}$, which has rank $d$, and let $X$ be symmetric of rank at most $d$ with $X_{uv}=M_{uv}$ for all $u\in\mathcal R$ and $v\ne u$. First, $X_{\mathcal R,\mathcal V\setminus\mathcal R}=M_{\mathcal R,\mathcal V\setminus\mathcal R}=B\Pi_{\mathcal V\setminus\mathcal R}^\top$ has rank $d$ because $\operatorname{rank}\Pi_{\mathcal V\setminus\mathcal R}=d$, and its column space is $\operatorname{col}(B)$. Since $X_{\mathcal R,\cdot}$ has rank at most $d$ and contains these columns, $\operatorname{col}(X_{\mathcal R,\cdot})=\operatorname{col}(B)$. Second, for $u\in\mathcal R$ the column $z=X_{\mathcal R,u}-M_{\mathcal R,u}$ lies in $\operatorname{col}(B)$ and vanishes off coordinate $u$. Writing $z=B\alpha$, the rows of $B$ other than $u$ give $\Pi_{\mathcal R\setminus\{u\}}C\alpha=0$, and $\operatorname{rank}\Pi_{\mathcal R\setminus\{u\}}=d$ with $C$ nonsingular forces $\alpha=0$, so $X_{uu}=M_{uu}$. Hence $X_{\mathcal R,\cdot}=M_{\mathcal R,\cdot}$, and since $M_{\mathcal R\mathcal R}=\Pi_{\mathcal R}C\Pi_{\mathcal R}^\top$ has rank $d$, Fact 2 gives $X=M$. A zero-residual point $(\Pi',C')$ of Problem~\eqref{Opt:fit} yields such an $X=\Pi'C'\Pi'^\top$. If the memberships are in general position, every $d$ of them are independent, which gives both rank conditions when $r\ge d+1$ and $n-r\ge d$. For pure memberships with $\theta_i>0$, $\operatorname{rank}\Pi_{\mathcal I}$ equals the number of communities represented in $\mathcal I$, which gives the stated characterization.

\emph{(ii), necessity.} Let $r\le d$ with $\operatorname{rank}\Pi_{\mathcal R}=r$ and an interior representative, and let $B_0=\Pi_{\mathcal R}C$, of full row rank $r$. The map $E\mapsto(\Pi_{\mathcal R}+E)C(\Pi_{\mathcal R}+E)^\top$ from $\mathbb R^{r\times d}$ to the symmetric $r\times r$ matrices has differential $E\mapsto EB_0^\top+B_0E^\top$ at $E=0$, which is onto: for a symmetric target $Z$, the choice $E=\tfrac12Z(B_0^{+})^\top$ gives $EB_0^\top=\tfrac12Z(B_0B_0^{+})^\top=\tfrac12Z$ because $B_0B_0^{+}=I_r$. By the implicit function theorem, for every sufficiently small diagonal matrix $\Delta$ there is a small $E$ with $(\Pi_{\mathcal R}+E)C(\Pi_{\mathcal R}+E)^\top=M_{\mathcal R\mathcal R}+\Delta$. Set $\Pi'_{\mathcal R}=\Pi_{\mathcal R}+E$ and, for $v\notin\mathcal R$, let $\pi'_v$ be the solution of $\Pi'_{\mathcal R}C\pi=M_{\mathcal R,v}$ closest to $\pi_v$, which exists because $\Pi'_{\mathcal R}C$ has full row rank and lies within $\mathcal O(\|E\|)$ of $\pi_v$. For small $\Delta\ne0$ all memberships stay in $\Delta_d$, and $M'=\Pi'C\Pi'^\top$ agrees with $M$ on every revealed entry while $M'_{uu}\ne M_{uu}$ for some $u\in\mathcal R$.

\emph{(iii), necessity.} Fact 1 with $\mathcal S=\mathcal R$: the revealed block has $\binom r2$ entries, and $\binom r2<\dim\mathcal M_{\mathcal R}$ whenever $r<r_0(d)$, since $\binom r2\ge rd-\binom d2$ is equivalent to $r^2-(2d+1)r+d^2-d\ge0$, whose larger root is $(2d+1+\sqrt{8d+1})/2$, and $\binom r2<\binom{r+1}2$ always.

\emph{(iii), sufficiency.} The off-diagonal entries of $M_{\mathcal R\mathcal R}$ form the row pattern of (ii) inside the block, with revealed rows $\mathcal R_1$ and unrevealed nodes $\mathcal R\setminus\mathcal R_1\supseteq\mathcal R_2$, so (ii) gives $X_{\mathcal R\mathcal R}=M_{\mathcal R\mathcal R}$ for every symmetric $X$ of rank at most $d$ with the same revealed entries. Let $v\notin\mathcal R$ be measured against $R_v$. As in Fact 2, $X$ has rank $d$ and its row space is that of $X_{\mathcal R,\cdot}$, so $X_{v,\mathcal R}$ lies in the row space of $X_{\mathcal R\mathcal R}=M_{\mathcal R\mathcal R}$, a $d$-dimensional subspace of $\mathbb R^{r}$. The coordinate projection of this subspace onto $R_v$ is injective if and only if $(M_{\mathcal R\mathcal R})_{\cdot,R_v}=\Pi_{\mathcal R}C\Pi_{R_v}^\top$ has rank $d$, that is, $\operatorname{rank}\Pi_{R_v}=d$, and then the revealed coordinates $X_{v,R_v}=M_{v,R_v}$ determine $X_{v,\mathcal R}=M_{v,\mathcal R}$. Once this holds for every $v\notin\mathcal R$, $X_{\mathcal R,\cdot}=M_{\mathcal R,\cdot}$ by symmetry, and Fact 2 gives $X=M$. \hfill$\square$

\paragraph{(H) Proofs of Theorem~\ref{thm:regret} and Corollary~\ref{cor:rows}, and a deterministic instance of Assumption~\ref{ass:confidence}.}
\emph{Theorem~\ref{thm:regret}.} We prove the stronger subset form. For any nonempty $I\subseteq\{0,\ldots,T-1\}$ with $m=|I|$, replace $T$ in the denominator of the theorem by $m$ and its error average by $m^{-1}\sum_{t\in I}(\Delta_t(v^{(t+1)})+\Delta_t(v^*))$, retaining $\gamma_T$. If $v^*\in\mathcal V^{(t)}$ for some $t<T$ then $r_T=0$. Otherwise Lemma~\ref{lem:optimism} applies in every round $t\in I$, and the simple regret is at most the average over $I$ of its right-hand side. It remains to bound $\frac1m\sum_{t\in I}s^\circ_t(v^{(t+1)})$. By Lemma~\ref{lem:reduction}, $s^\circ_t(v)^2-\sigma_b^2=\phi_v^\top\Xi_t\phi_v$, where $\Xi_t$ is the posterior covariance of a $d_+$-dimensional Bayesian linear model with prior covariance $D$ and per-node bulk variance $\sigma_b^2$ on the evaluated set. Since $\Xi_t\preceq D$, one has $\phi_v^\top\Xi_t\phi_v\le v_{\max}$, hence $\phi^\top\Xi_t\phi\le\max(\sigma_b^2,v_{\max})\min(1,\phi^\top\Xi_t\phi/\sigma_b^2)$, and the elliptical potential lemma~\citep{abbasi2011improved} applied to the actual evaluation sequence gives $\sum_{t<T}\min\bigl(1,\phi_{v^{(t+1)}}^\top\Xi_t\phi_{v^{(t+1)}}/\sigma_b^2\bigr)\le2\log\det\bigl(I+D^{1/2}G_SD^{1/2}/\sigma_b^2\bigr)\le2\gamma_T$, where the last step uses $\log\det(I+X)\le d_+\log(1+\operatorname{tr}X/d_+)$ and $\operatorname{tr}(D^{1/2}G_SD^{1/2})=\sum_{s\in S}\phi_s^\top D\phi_s\le(T+N_0)\,v_{\max}$. The sum over $I$ is a partial sum of nonnegative terms and obeys the same bound, so $\sum_{t\in I}\phi_{v^{(t+1)}}^\top\Xi_t\phi_{v^{(t+1)}}\le c_1\gamma_T$. By $\sqrt{a+b}\le\sqrt a+\sqrt b$ and the Cauchy--Schwarz inequality, $\frac1m\sum_{t\in I}s^\circ_t(v^{(t+1)})\le\sigma_b+\sqrt{c_1\gamma_T/m}$. \hfill$\square$

\emph{Corollary~\ref{cor:rows}.} For $t\ge t_{\mathrm{id}}$ the surrogate is identified and the kernel parameters and conditioning agree with the reference, so the zero-residual point of Problem~\eqref{Opt:fit} is the reference surrogate, and the fitted features are the reference features up to an orthogonal transformation within each eigenspace of $\tilde L_\zeta$, which leaves the kernel, the posterior, and the scores unchanged. Hence $\Delta_t=0$ for $t\ge t_{\mathrm{id}}$, and the subset form just proved with $I=\{t_{\mathrm{id}},\dots,T-1\}$ gives the bound. Under Algorithm~\ref{Alg:global_bo} the rows of the $N_0$ initial nodes are revealed at the start and $Q_R$ rows before each decision, so $N_0+(t+1)Q_R$ rows are revealed at the decision of round $t$. In general position any $d+1$ distinct rows satisfy Proposition~\ref{prop:coverage}(ii) when $n\ge2d+1$, so identification occurs no later than $(\lceil(d+1-N_0)_+/Q_R\rceil-1)_+$ decisions. Later observations preserve uniqueness. For interior representatives, $d$ or fewer rows leave nonunique completions, giving the matching threshold in general. \hfill$\square$

\emph{Deterministic instance of Assumption~\ref{ass:confidence}.} Let $y=y_s+y_b$ as in Assumption~\ref{ass:split}, and write the posterior mean as $\mu^\circ_t(v)=a_t(v)^\top\mathbf y_S$ with weights $a_t(v)=\tilde U_S(\sigma_b^2D^{-1}+G_S)^{-1}\phi_v$, by Lemma~\ref{lem:reduction} and the push-through identity. On $\mathcal H(K_s)$ the functional $f\mapsto f(v)-a_t(v)^\top f_S$ is bounded, with squared norm $K_s(v,v)-2a_t(v)^\top K_{s,Sv}+a_t(v)^\top K_{s,SS}a_t(v)=\phi_v^\top\Xi_t\phi_v-\sigma_b^2\|a_t(v)\|_2^2\le\phi_v^\top\Xi_t\phi_v$, the kernel ridge regression identity with regularization $\sigma_b^2$. Hence $|y_s(v)-a_t(v)^\top\mathbf y_{s,S}|\le B_s(\phi_v^\top\Xi_t\phi_v)^{1/2}$, while $|y_b(v)-a_t(v)^\top\mathbf y_{b,S}|\le\eta(1+\|a_t(v)\|_1)$. Therefore
\[
|y(v)-\mu^\circ_t(v)|\ \le\ B_s\,s^\circ_t(v)+\eta\bigl(1+\|a_t(v)\|_1\bigr)\qquad\text{for every }v\text{ and }t,
\]
deterministically, so Assumption~\ref{ass:confidence} holds with $\kappa=B_s+\eta(1+\alpha)/\sigma_b$ whenever the posterior weights satisfy $\max_{t,v}\|a_t(v)\|_1\le\alpha$, because $s^\circ_t\ge\sigma_b$.

\paragraph{(I) Errors in variables at evaluated nodes.}
With $w_t=(\sigma_b^2D^{-1}+G_S)^{-1}\tilde U_S^\top\mathbf y_S$, a perturbation $\tilde U_S\mapsto\tilde U_S+E$ of the evaluated nodes' features changes the shared coefficient at first order by $\delta w_t=(\sigma_b^2D^{-1}+G_S)^{-1}\bigl[E^\top\mathbf y_S-(E^\top\tilde U_S+\tilde U_S^\top E)w_t\bigr]$, and therefore every prediction by $\delta\mu_t(v)=\phi_v^\top\delta w_t$. Making the features of evaluated nodes accurate is thus a global correction, which is why Algorithm~\ref{Alg:structure} reveals the rows of evaluated nodes first.

\paragraph{(J) Why the unsplit smoothness constant is of order $\sqrt n$.}
A single RKHS bound $\|y\|_{\mathcal H(K)}\le B$ for the full-rank kernel $K=(\beta I+L_\zeta)^{-1}$ contains $\beta\|y\|_2^2$ by the identity of (A), so $B^2\ge\beta\|y\|_2^2$. For an objective standardized to unit variance over the nodes, $\|y\|_2^2=n$ and hence $B\ge\sqrt{\beta n}$. Any regret bound proportional to $B$, and any confidence width $Bs_t(v)$ with $s_t(v)\ge(\beta+2)^{-1/2}$, is then at least of order $\sqrt{\beta n/(\beta+2)}$, which is $9.5$ at $n=1000$ and $\beta=0.2$, while the range of such an objective is about $6$. Assumption~\ref{ass:split} removes the $\sqrt n$ by bounding the bulk component in the supremum norm instead of through the RKHS norm.

\paragraph{(K) The scalar $\delta$, the knowledge-gradient form, and Lemma~\ref{lem:leader}.}
\begin{proposition}[Value of a pair query]
\label{prop:kg}
In the working model with the global geometry and the coefficient state held fixed and the memberships independent across nodes, let $Z_w=U_t(w)+c_w^\top(\pi_w-\hat\pi_w)$ be the linearized scores of the unevaluated nodes and $\mathcal L(\mathcal H)=\mathbb E[\max_wZ_w\mid\mathcal H]-\max_w\mathbb E[Z_w\mid\mathcal H]$ the expected loss of selecting by the current scores given the information $\mathcal H$. A query of $v$ along $g$ with answer $a$ reduces this loss in expectation by
\[
\mathrm{KG}_t(v,g)=\mathcal L(\mathcal H)-\mathbb E_a[\mathcal L(\mathcal H,a)\mid\mathcal H]=\delta(v,g)\,f\Bigl(-\frac{\mathrm{gap}_t(v)}{\delta(v,g)}\Bigr),\qquad f(z)=z\Phi(z)+\varphi(z),
\]
where $\mathrm{gap}_t(v)=|U_t(v)-\max_{w\ne v}U_t(w)|$ and $\Phi,\varphi$ are the standard normal distribution and density. The value increases with $\delta$, decreases with the gap, and vanishes when $\delta=0$.
\end{proposition}
Condition $\pi_v\sim\mathcal N(\hat\pi_v,\Sigma)$ on $a=\langle g,\pi_v\rangle+e$ with $e\sim\mathcal N(0,\sigma^2)$, where the exact observations of this paper correspond to $\sigma^2\to0$. The predictive law of $a$ is $\mathcal N(\langle g,\hat\pi_v\rangle,\,g^\top\Sigma g+\sigma^2)$, the posterior mean is $\hat\pi_v+\Sigma g\,(a-\langle g,\hat\pi_v\rangle)/(g^\top\Sigma g+\sigma^2)$, and in the exact-observation limit the posterior covariance is Eq.~\eqref{Equ:rankone}. Hence $c^\top\hat\pi_v'-c^\top\hat\pi_v$ is Gaussian with standard deviation $|c^\top\Sigma g|/\sqrt{g^\top\Sigma g+\sigma^2}=\delta$, and $c^\top\Sigma c-c^\top\Sigma'c=\delta^2$, which is Eq.~\eqref{Equ:delta}. \emph{Proposition~\ref{prop:kg}.} By the tower property, $\mathbb E_a[\mathbb E[\max_wZ_w\mid\mathcal H,a]\mid\mathcal H]=\mathbb E[\max_wZ_w\mid\mathcal H]$, so $\mathcal L(\mathcal H)-\mathbb E_a[\mathcal L(\mathcal H,a)\mid\mathcal H]=\mathbb E_a[\max_w\mathbb E[Z_w\mid\mathcal H,a]\mid\mathcal H]-\max_w\mathbb E[Z_w\mid\mathcal H]$, the expected increase of the maximal conditional score mean. Before the query the conditional means are $U_t(w)$. After it, the mean of $Z_v$ is $U_t(v)+\delta Z$ with $Z$ standard normal under the predictive law of $a$, by the computation above, and the other means are unchanged because the target memberships are independent in the working model. With $A=U_t(v)$ and $B=\max_{w\ne v}U_t(w)$, direct integration gives $\mathbb E\max(A+\delta Z,B)-\max(A,B)=\delta\varphi(|A-B|/\delta)-|A-B|\Phi(-|A-B|/\delta)=\delta f(-|A-B|/\delta)$. Its derivative in $\delta$ is $\varphi(|A-B|/\delta)>0$, its derivative in the gap is $-\Phi(-|A-B|/\delta)<0$, and it tends to $0$ as $\delta\to0$. For a plan of $m$ queries with $c_v$ held fixed, the joint Gaussian update gives the change of the mean of $Z_v$ the variance $c_v^\top(\Sigma_v-\Sigma_v^{(m)})c_v=\delta_m(v)^2$, and the same integral gives the value of the plan, which Eq.~\eqref{Equ:kg} divides by $m$.

\emph{Lemma~\ref{lem:leader}.} A probe of $v$ along a landmark direction $g$ with $g^\top\Sigma_vg>0$ lowers the rank of $\Sigma_v$ by one, by Eq.~\eqref{Equ:rankone}, and Algorithm~\ref{Alg:pair} probes only such directions. Under the spanning assumption, every unpinned unevaluated node has an eligible query until it is pinned, which takes at most $d$ probes per node. The initial pin step uses at most $d$ probes for $v^{(t)}$. Subsequently, each query reduces the sum of the remaining covariance ranks over the $m_t$ unpinned unevaluated nodes by one, so at most $dm_t$ further queries pin them all. Nodes already pinned remain exact under the fixed exact geometry. Thus $Q_E\ge d(m_t+1)$ suffices to make every unevaluated node's feature exact, including the selected node and the optimum if it is still unevaluated. No confidence-based candidate exclusion is used. \hfill$\square$

\paragraph{(L) Proof of Corollary~\ref{cor:pairs}.}
With the evaluated nodes pinned, $\Sigma_v=0$ there and the moment adjustment of Eq.~\eqref{Equ:eiv} is void, so exact conditioning by Lemma~\ref{lem:reduction} with the reference prior $D$ and exact evaluated features gives the reference coefficient state, so Lemma~\ref{lem:optimism} gives $\Delta_t(v)\le\Lambda_te_t(v)$ at every unevaluated $v$, and Theorem~\ref{thm:regret} gives the display. The argument uses neither $A_{\mathrm{off}}\in\mathcal C_d$ nor the identification of the skeleton, so for any graph the same display holds with $e_t$ the distance to the reference features. Under an identified skeleton with exact global quantities, the spanning conditions, and the budget $Q_E\ge d(m_t+1)$ of Lemma~\ref{lem:leader}, every unevaluated node is pinned at decision time, including $v^*$ and $v^{(t+1)}$, so $e_t(v^{(t+1)})=e_t(v^*)=0$. \hfill$\square$

\paragraph{(M) Decision error under an estimated geometry.}
\begin{lemma}[Decision error under an estimated geometry]
\label{lem:optimism}
Under Assumption~\ref{ass:confidence}, if $v^*$ is unevaluated and $v^+=v^{(t+1)}$ maximizes $U_t$, Eq.~\eqref{Equ:perround} holds. If the algorithm and reference use the same bulk variance and coefficient state $(w_t,\Xi_t)$, then $\Delta_t(v)\le\Lambda_t e_t(v)$, where $e_t(v)=\|\hat\phi_v-\phi_v\|_2$ in a common feature basis and $\Lambda_t=\|w_t\|_2+\kappa\|\Xi_t\|_2^{1/2}$. Exact evaluated features, the reference prior, and exact coefficient conditioning suffice for this state agreement.
\end{lemma}
On the event of Assumption~\ref{ass:confidence}, $y(v^*)\le U^\circ_t(v^*)\le U_t(v^*)+\Delta_t(v^*)\le U_t(v^+)+\Delta_t(v^*)\le U^\circ_t(v^+)+\Delta_t(v^+)+\Delta_t(v^*)$, where the third step is the maximality of $v^+$ under $U_t$, and $y(v^+)\ge\mu^\circ_t(v^+)-\kappa s^\circ_t(v^+)=U^\circ_t(v^+)-2\kappa s^\circ_t(v^+)$. Subtracting the two displays gives Eq.~\eqref{Equ:perround}. For the second claim, when the coefficient state is the reference state both scores are the same function $U(\phi)=\phi^\top w_t+\kappa(\sigma_b^2+\phi^\top\Xi_t\phi)^{1/2}$, evaluated at $\hat\phi_v$ and at $\phi_v$. Its gradient is $w_t+\kappa\,\Xi_t\phi/(\sigma_b^2+\phi^\top\Xi_t\phi)^{1/2}$, and $\|\Xi_t\phi\|_2\le\|\Xi_t\|_2^{1/2}(\phi^\top\Xi_t\phi)^{1/2}$, so $U$ is $\Lambda_t$-Lipschitz and $\Delta_t(v)\le\Lambda_te_t(v)$ by the mean value inequality. \hfill$\square$

\section{Pair Queries: Landmark Skeleton and Knowledge-Gradient Probes}
\label{app:pair}
This appendix completes the pair-query design of Section~\ref{Subsec:edge design}: the cold start that creates landmarks, the greedy KG($*$) procedure, and the lateration lemma that bounds its budget. For a linear functional $c^\top\pi_v$ the rank-one update of Eq.~\eqref{Equ:rankone} reads
\begin{equation}
\label{Equ:rankone-c}
c^\top\Sigma_v'c=c^\top\Sigma_vc-\frac{(c^\top\Sigma_vg)^2}{g^\top\Sigma_vg},
\end{equation}
a query along an already measured direction has $g^\top\Sigma_vg=0$ and is skipped, and $d$ queries with spanning directions pin $v$ conditional on the fixed landmark geometry.

\paragraph{Cold start: a landmark skeleton.}
Nothing can start without landmarks, and the averages that every feature depends on, namely $\Sigma_0$, the degree vector $h=C\Pi^\top\mathbf 1$, and the orthonormalization of Lemma~\ref{lem:exact-solve}, must be estimated from nodes with known memberships. The experiments create the landmarks from the block among the $N_0$ initial nodes, $\binom{N_0}2$ pairs, which identifies the landmark memberships once $N_0\ge2d+1$ in general position and needs $N_0\ge r_0(d)$ (Proposition~\ref{prop:coverage}(iii)). A node that has not been probed sits at the mean landmark membership until its first query, which is why the queries of a round precede its selection. An alternative skeleton probes every node against $r$ uniformly drawn landmarks at cost $r(n-1)$ entries, reproduces the row pattern, and identifies the surrogate once $r\ge d+1$ in general position (Proposition~\ref{prop:coverage}(ii)).

\paragraph{Greedy allocation.}
The eligible set is
\begin{equation}
\label{Equ:pair-candidates}
\mathcal N_t=\{v\notin\mathcal V^{(t)}:\exists u\in\mathcal R,\ (v,u)\notin\Omega^{(t)},\ g_u^\top\Sigma_vg_u>0\}.
\end{equation}
Algorithm~\ref{Alg:pair} first pins the node evaluated last, forms the scores and the eligible set $\mathcal N_t$ of Eq.~\eqref{Equ:pair-candidates}, and then spends the remaining budget greedily by the KG($*$) score of Eq.~\eqref{Equ:kg}, updating eligibility and re-optimizing the plan after every answer. All unevaluated, unpinned nodes with an available informative landmark pair are eligible, and there is no confidence-radius threshold. Budget that is left when no eligible query remains is carried to the next round. Since the gains are not submodular in general (Appendix~\ref{app:derivations}(F)), the plan is re-optimized after every answer, and Lemma~\ref{lem:leader} gives a budget that pins every unevaluated node, including the selected one.

\begin{algorithm}[t]
\caption{Entry Acquisition by the Knowledge Gradient KG($*$) (one round)}
\label{Alg:pair}
\textbf{Input:} observations $\Omega$, landmarks $\mathcal R$, budget $Q_E$, fit $(\Pi,C)$, $\{\Sigma_v\}$, GP state $(w_t,\Xi_t)$, the node $v^{(t)}$ evaluated last\\
\textbf{Output:} $\Omega^{(t+1)}$ and the node $v^{(t+1)}$ to evaluate
\begin{algorithmic}[1]
\STATE $q\gets0$. If $v^{(t)}$ is not pinned, probe it along up to $d$ spanning landmark directions within the pin share $\rho_{\mathrm{pin}}$ of $Q_E$, add the number of probes to $q$, and recompute $(w_t,\Xi_t)$
\hfill$\triangleright$ exact features at evaluated nodes
\STATE compute $U_t$ by Eq.~\eqref{Equ:ucb-struct} over unevaluated nodes, and the eligible set $\mathcal N_t$ by Eq.~\eqref{Equ:pair-candidates}
\WHILE{$q<Q_E$ and $\mathcal N_t\ne\varnothing$}
  \STATE for each $v\in\mathcal N_t$: $u_v\gets\arg\max\{\delta(v,g_u):\ u\in\mathcal R,\ (v,u)\notin\Omega,\ g_u^\top\Sigma_vg_u>0\}$ with $c=c_v$, and $\mathrm{KG}^\ast_t(v)$ by Eq.~\eqref{Equ:kg}
  \STATE $v^\star\gets\arg\max_{v\in\mathcal N_t}\mathrm{KG}^\ast_t(v)$ with ties broken at random, and reveal $A_{v^\star u_{v^\star}}$
  \STATE update $\Omega$, $\hat\pi_{v^\star}$, $\Sigma_{v^\star}$, $\hat\phi_{v^\star}$, $c_{v^\star}$ and $U_t(v^\star)$; refresh $\mathcal N_t$ and set $q\gets q+1$
  \hfill$\triangleright$ $\mathcal O(|\mathcal R|d^3)$ for the probed node, all other nodes unchanged
\ENDWHILE
\STATE $v^{(t+1)}\gets\arg\max_vU_t(v)$ over unevaluated nodes, and carry $Q_E-q$ to the next round
\end{algorithmic}
\end{algorithm}

\begin{lemma}[Lateration over all unevaluated nodes]
\label{lem:leader}
Let $A_{\mathrm{off}}\in\mathcal C_d$ with an identified landmark skeleton whose global quantities are exact and held fixed during the round. Suppose that for every unpinned unevaluated node, and for the node evaluated last if it is unpinned, the observed and available landmark directions together span its $d$ membership coordinates. Run Algorithm~\ref{Alg:pair} with budget $Q_E$ and a pin share that admits $d$ probes, and let $m_t$ be the number of all unpinned unevaluated nodes at the start of the round. If $Q_E\ge d(m_t+1)$, then every unevaluated node is pinned by the end of the round, so that $e_t(v^{(t+1)})=0$ and, if the optimum is unevaluated, $e_t(v^*)=0$ at decision time.
\end{lemma}
The proof is in paragraph (K) of Appendix~\ref{app:derivations}. Each step of Algorithm~\ref{Alg:pair} costs $\mathcal O(|\mathcal R|d^3)$ for the probed node and $\mathcal O(n)$ for the maxima, and a round costs $\mathcal O(n|\mathcal R|d^3+Q_E(|\mathcal R|d^3+n))$ after the refresh of Algorithm~\ref{Alg:kernel_learning}.

\section{Initialization and Refit Schedule}
\label{app:coldstart}
\paragraph{Solver.}
Problem~\eqref{Opt:fit} is solved by projected alternating least squares. Given the core and the other memberships, the objective is a convex quadratic in each $\pi_i$, and under row queries every unrevealed node solves the same shared-design problem $\min_{\pi\in\Delta_d}\|G_{\mathcal R}\pi-A_{\mathcal R,v}\|_2^2$ with $G_{\mathcal R}=\Pi_{\mathcal R}C$, for all nodes at once. Given the memberships, the core solves a box-constrained least-squares problem over symmetric $C\in[0,1]^{d\times d}$. Each subproblem is solved by accelerated projected gradient with Euclidean projection onto $\Delta_d$ or onto the symmetric box, a block update is kept only if the residual does not rise, and one sweep updates the revealed memberships, the unrevealed memberships, and the core in turn, at cost $\mathcal O(|\Omega|d+nd^2)$ under row queries and $\mathcal O(|\Omega|d^2+nd^2)$ under pair queries, where each node carries its own design. The sub-simplex $\Delta_d$ is the simplex over $d+1$ communities with the weight of a null community, which interacts with no one, dropped: a membership has $d$ free coordinates, $d$ measurements pin a node, and the class contains the degree-corrected block memberships $\pi_i=\theta_ie_{k(i)}$ with $\theta_i\in[0,1]$, which the simplex $\{b\ge0,\ \mathbf 1^\top b=1\}$ does not.

\paragraph{Structured initialization.}
Problem~\eqref{Opt:fit} is nonconvex, and the quality of a local solver depends on its starting point. Row observations supply a natural one. The revealed strip $A_{\mathcal R,\cdot}$ is an $r\times n$ matrix that is fully observed except for its $r$ self-pairs, and under the class it equals $\Pi_{\mathcal R}C\Pi^\top$ off those pairs. We compute a rank-$d$ factorization of the strip in two ways, by a truncated SVD whose factors are rescaled and projected onto $\Delta_d$, and by successive projection, which selects $d$ columns of the strip as approximate pure nodes. Each candidate initializes $\Pi$. The core $C$ is then obtained from the box-constrained least-squares problem, a fixed number of sweeps (12 in the experiments) is run from each candidate, and the candidate with the smaller residual is kept.

\paragraph{Warm starts and periodic refits.}
After every reveal the fit is updated by a few sweeps (four in the experiments) started from the previous solution, which is cheap and tracks the data when the pattern changes little. A warm-started iterate can nevertheless settle in a poor local minimum, especially during the first rounds, when each new row changes the strip substantially. Every $k_{\mathrm{refit}}$ revealing rounds the structured initialization is therefore recomputed on the current strip, the fit is run from it, and the better of the fresh and the warm-started solutions in terms of the residual is kept, with the warm-started solution retained on a tie. The experiments use $k_{\mathrm{refit}}=10$, chosen on the design seeds.

\paragraph{Calibration of the shared averages.}
The reference covariance $\Sigma_0$, the degree vector $h=C\Pi^\top\mathbf 1$, and the orthonormalization $\sum_uM_uM_u^\top$ that every feature depends on are averages over nodes. Under row queries all $n$ memberships are estimated from the first round on, so these averages are computed over all nodes. Their accuracy depends on the fitted memberships. The uniform component of the row policy provides coverage beyond the adaptively evaluated nodes.

\section{Additional Experimental Results}
\label{app:exp}
\begingroup
{This appendix gives the protocols and the complete results behind the three questions of Section~\ref{sec:experiments}. Results from different initializations or budget schedules are never pooled.}

\subsection{Protocols and Structural Information Accounting}
\label{app:exp-protocols}
\begin{table}[t]
\centering\small
\setlength{\tabcolsep}{4pt}
\caption{Protocols used in Section~\ref{sec:experiments}. $T$ counts evaluations after initialization, and the row totals exclude the initial rows. BioPlex targets share the indicated protocol. The full-information study is a separate campaign with the horizons below, and ESOL uses a different initialization from the main row study.}
\label{tab:exp-protocols}
\begin{tabular}{lrrrrl}
\toprule
Study / instance & $n$ & $N_0$ & $T$ & $d$ & Additional structure\\\midrule
Main: dense (two objectives) & 1,000 & 8 & 50 & 16 & $Q_R=2$, $TQ_R=100$ rows\\
Main: ESOL & 1,117 & 8 & 60 & 16 & $Q_R=2$, $TQ_R=120$ rows\\
Main: Chameleon & 1,223 & 8 & 100 & 16 & $Q_R=5$, $TQ_R=500$ rows\\
Main: BioPlex 293T & 13,167 & 20 & 100 & 16 & $Q_R=10$, $TQ_R=1{,}000$ rows\\
Main: BioPlex HCT116 & 9,609 & 20 & 100 & 16 & $Q_R=10$, $TQ_R=1{,}000$ rows\\
Main: HCT116 / own line & 9,494 & 20 & 100 & 16 & $Q_R=10$, $TQ_R=1{,}000$ rows\\
\midrule
Sparse tasks (Table~\ref{tab:exp-additional-rows}) & 2,000 / 5,000 & 8 & 100 & 16 & 400 / 1,000 total rows\\
Scale ablation: 293T / DCSBM & 13,167 / 5,000 & 20 / 8 & 100 & 16 & 250 / 1,000 total rows\\
\midrule
Full information: BioPlex 293T & 13,167 & 20 & 100 & 4--32 & Full graph\\
Full information: DCSBM & $\approx5{,}000$ & 8 & 100 & 4--32 & Full graph\\
Full information: ESOL & 1,117 & 32 & 100 & 4--32 & Full graph, score at 25\\
\midrule
ESOL per-round pair study & 1,117 & 32 & 200 & 32 & $Q_E=2,4,8,16,32$\\
ESOL row-rule study & 1,117 & 32 & 200 & 16 & $Q_R=.1,.25,.5,1,2$\\
\bottomrule
\end{tabular}
\end{table}
\paragraph{Budget conventions.}
The row totals of Table~\ref{tab:exp-protocols} are cumulative additional rows, excluding initialization. Revealing $r$ distinct rows discloses $r(n-1)-r(r-1)/2$ distinct off-diagonal entries. Both nonzero and zero answers count. Entries shared by two queried rows count once in the observed-entry set, while both row queries are charged. Pair queries target previously unobserved unordered pairs. In the main row study, exactly $Q_R$ new rows are bought before each post-initialization evaluation, $TQ_R$ in total, and local graph BO has the same allowance and may use fewer rows. In the pair study, all three acquisition methods receive exactly $Q_E$ new pairs before every post-initialization evaluation, including the first, following the common 496-pair initial block.

The row-rule study of Appendix~\ref{app:exp-rows} uses $Q_R$ as an average acquisition rate. The integer allocation is $q_t=\lfloor tQ_R\rfloor-\lfloor(t-1)Q_R\rfloor$, so cumulative additional rows are $\lfloor tQ_R\rfloor$. In particular, $Q_R=0.1$ buys one complete row every ten evaluations (20 rows by $T=200$), not a fraction of a row. This is a fixed schedule with zero-query rounds, distinct from acquiring a fixed positive integer number of rows at every iteration.

\paragraph{Comparison axes.}
{Local graph BO~\citep{wan2023bayesian} receives the same cumulative row allowance as Ours and builds its neighborhoods from its own revealed rows on Chameleon and BioPlex. On the dense and ESOL affinities it uses a precomputed $k$NN view of the full graph, which conveys structural information beyond the allowance and is disclosed in Section~\ref{sec:exp-setup}. In the pair study it is adapted to the pair oracle, its local context is built from the revealed entries, and its restarts are part of the adaptation. Full-graph references receive all structural entries but still query objectives sequentially.}

\paragraph{Statistics.}
{The main comparisons and the full-information study use 20 evaluation seeds, the observation-pattern study and the scale ablation use 10. The design seeds, on which every configuration was chosen, are disjoint from the evaluation seeds. Seeds and initial nodes are shared within each protocol. SEM is the sample standard deviation divided by the square root of the seed count, and uncertainty bands are pointwise. For the pair study we report percentile bootstrap intervals from 10,000 resamples of paired seeds, at the nominal 95\% level and at 99\% for a Bonferroni adjustment across the five budgets within each comparator family. No curve is smoothed, and every displayed cell has all its seeds.}

\subsection{Budget Sensitivity and Additional Graphs}
\label{app:exp-additional}
\paragraph{Row-query trajectories of the main campaign.} Figure~\ref{fig:exp-dense-rows} gives the trajectories of the campaign of Table~\ref{tab:results} on the dense in-class instance and on ESOL next to the BioPlex panel of Figure~\ref{fig:rows-main}. Both dense objectives remain in Table~\ref{tab:results}, and the class-prior objective provides a controlled setting in which the surrogate family matches the graph generator: it is drawn from the class kernel of the generating matrix, so the class fitted to 108 rows already matches the full-graph references within seed variability.
\begin{figure}[t]
\centering
\includegraphics[width=\linewidth]{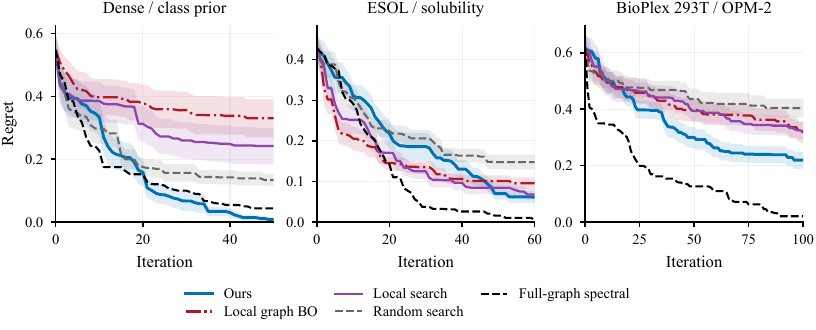}
\caption{Row-query trajectories of the campaign of Table~\ref{tab:results}: the dense in-class instance with the class-prior objective ($n=1{,}000$, $N_0=8$, $T=50$, $Q_R=2$), ESOL ($N_0=8$, $T=60$, $Q_R=2$), and BioPlex 293T with the OPM-2 objective ($N_0=20$, $T=100$, $Q_R=10$), all at $d=16$. Curves show mean normalized simple regret with pointwise $\pm1$ SEM bands over 20 paired seeds, and the dashed spectral reference sees the entire graph. On the dense and ESOL affinities local graph BO and local search use a precomputed $k$NN view of the full graph. The BioPlex panel is Figure~\ref{fig:rows-main}.}
\label{fig:exp-dense-rows}
\end{figure}

Figure~\ref{fig:exp-row-budgets} reports every archived row budget of Ours on the six BioPlex targets, with local graph BO at the same allowances. The $1{,}000$-row protocol of Table~\ref{tab:results} lies inside this range. The response to the budget is not monotone at every step, since each fit changes the evaluation path.

\begin{figure}[t]
\centering
\includegraphics[width=\linewidth]{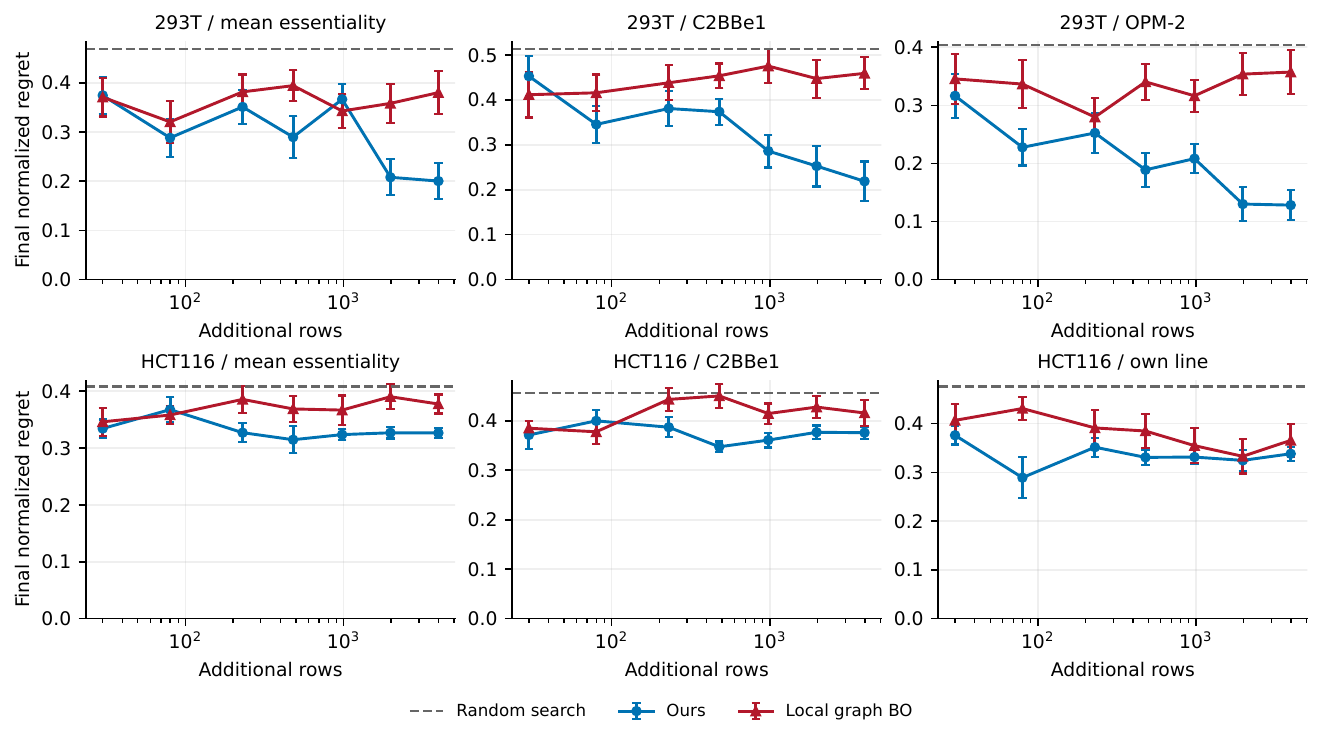}
\caption{BioPlex row-budget sensitivity, final normalized regret after $T=100$ evaluations following $N_0=20$. Additional rows are the rows revealed after initialization over all $T$ rounds, spent at a constant rate, in a campaign separate from Table~\ref{tab:results}. Error bars are $\pm1$ SEM over 20 seeds. Every archived Ours row budget is shown, and local graph BO runs on the original binary graph with the same row allowance. Dashed lines show random-search means.}
\label{fig:exp-row-budgets}
\end{figure}

Table~\ref{tab:exp-additional-rows} extends the comparison with random search to sparse synthetic graphs. The DCSBM and latent-position examples have lower Ours mean regret than random search at the displayed budgets.
\begin{table}[t]
\centering\small
\setlength{\tabcolsep}{4pt}
\caption{Additional row tasks, final normalized regret (mean $\pm$ SEM, 20 seeds), $d=16$, $T=100$, and $N_0=8$. Rows are total revealed rows including the initial ones, spent at a constant rate over the rounds. Only random search was run on these instances, and the two full-graph columns have complete structural information.}
\label{tab:exp-additional-rows}
\begin{tabular}{lrcccc}
\toprule
Instance / objective & Rows & Ours & Random & Ours (full) & Full spectral\\\midrule
DCSBM $n\approx2{,}000$ / latent & 400 & $.154\pm.030$ & $.209\pm.027$ & $.094\pm.023$ & $.088\pm.022$\\
DCSBM $n\approx5{,}000$ / latent & 1000 & $.165\pm.027$ & $.245\pm.024$ & $.113\pm.023$ & $.077\pm.020$\\
DCSBM $n\approx5{,}000$ / class prior & 1000 & $.120\pm.028$ & $.341\pm.024$ & $.020\pm.016$ & $.047\pm.033$\\
Latent-position $n\approx5{,}000$ & 1000 & $.177\pm.017$ & $.212\pm.018$ & $.101\pm.014$ & $.099\pm.015$\\
\bottomrule
\end{tabular}
\end{table}

\paragraph{Initial structure and full-information references.}
Table~\ref{tab:exp-structure-references} complements Table~\ref{tab:results} at the same task-specific initialization and objective-evaluation horizon. Initial-only retains $\Omega_0$ throughout BO: function observations still update the posterior, and periodic kernel-hyperparameter fitting remains enabled. Its comparison with Ours measures the value of acquiring structure after initialization. Ours, full graph fits the same surrogate family to all entries, whereas the all-mode reference uses the true graph's untruncated spectral kernel. The latter is tuned on its own normalization and hyperparameter grid and serves as the untruncated reference.
\begin{table}[t]
\centering\small
\setlength{\tabcolsep}{4pt}
\caption{Additional representation controls for the main row study, normalized regret (mean $\pm$ SEM, 20 seeds). The task-specific $N_0$ and $T$ match Table~\ref{tab:exp-protocols}. Initial-only buys no structure after initialization, and the other columns receive the entire graph. The full-graph $d=16$ spectral reference is in Table~\ref{tab:results}. The all-mode column is measured under this protocol, at the same initialization, horizon and seeds as the other two columns.}
\label{tab:exp-structure-references}
\begin{tabular}{lccc}
\toprule
Instance / objective & Initial-only & \shortstack{Ours, full graph\\$d=16$} & \shortstack{Full graph\\spectral, all modes}\\\midrule
Dense / class prior & $.287\pm.056$ & $.048\pm.027$ & $.341\pm.061$\\
Dense / latent & $.247\pm.038$ & $.170\pm.026$ & $.244\pm.037$\\
ESOL / solubility & $.145\pm.022$ & $.005\pm.005$ & $.121\pm.013$\\
Chameleon / log traffic & $.106\pm.020$ & $.095\pm.006$ & $.055\pm.015$\\
293T / mean essentiality & $.401\pm.037$ & $.112\pm.031$ & $.272\pm.039$\\
293T / C2BBe1 & $.492\pm.035$ & $.162\pm.037$ & $.289\pm.042$\\
293T / OPM-2 & $.324\pm.037$ & $.126\pm.021$ & $.285\pm.037$\\
HCT116 / mean essentiality & $.374\pm.016$ & $.314\pm.006$ & $.344\pm.022$\\
HCT116 / C2BBe1 & $.421\pm.023$ & $.358\pm.021$ & $.364\pm.024$\\
HCT116 / own line (HCT116) & $.451\pm.024$ & $.309\pm.016$ & $.359\pm.023$\\
\bottomrule
\end{tabular}
\end{table}
{\paragraph{Representation and estimation gaps.}
Section~\ref{sec:exp-q2} reads Tables~\ref{tab:results} and~\ref{tab:exp-structure-references} as a decomposition at $d=16$. The estimation gap of a task is the regret of Ours on the revealed rows minus the regret of the same class fitted to the entire graph, and the representation gap is the latter minus the regret of the full-graph spectral reference at the same $d$. The estimation gaps are $-0.040$ and $-0.012$ on the dense class-prior and latent objectives, $-0.016$ on Chameleon, $0.056$ on ESOL, $0.192$, $0.064$, and $0.093$ on BioPlex 293T (mean essentiality, C2BBe1, OPM-2), and $0.009$, $-0.009$, and $0.050$ on HCT116. The representation gaps are $0.005$ and $0.000$ on the dense objectives, $0.018$ on Chameleon, $0.000$ on ESOL, $0.112$, $0.162$, and $0.105$ on 293T, and $-0.016$, $0.028$, and $0.011$ on HCT116. On the dense instance and Chameleon the partial fit is at or below the full fit, so what remains is representation. On ESOL and on 293T with mean essentiality estimation dominates, on 293T with C2BBe1 and OPM-2 the two gaps are comparable, and on HCT116 both are small. The two class fits share $d$, filter, initialization, and horizon, so the estimation gap isolates the revealed entries, and the full-graph spectral reference, tuned on its own normalization, makes the representation gap a comparison of two representations of the same graph at the same $d$.}

Check against the per-seed records that the dense/latent and ESOL cells of the middle column of Table~\ref{tab:exp-structure-references} are not copies of the full-graph spectral column of Table~\ref{tab:results}, since the two coincide to three decimals.

\subsection{Full-Information Representation Study}
\label{app:exp-e6}
\begin{figure}[t]
\centering
\includegraphics[width=\linewidth]{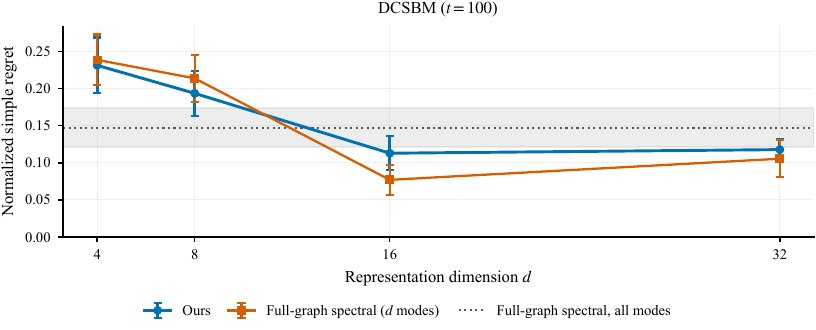}
\caption{DCSBM full-information resolution control, complementing Figure~\ref{fig:exp-resolution}. The graph generator has 16 communities and nominal $n=5{,}000$ before giant-component restriction, and the objective is latent, with $N_0=8$ and $T=100$. Error bars and the all-mode reference band are $\pm1$ SEM over 20 seeds. Both finite-dimensional arms improve up to $d=16$, and larger $d$ does not improve regret further. The all-mode kernel is tuned on its own grid.}
\label{fig:exp-dcsbm-resolution}
\end{figure}
\paragraph{Question and protocol.}
The full-information study asks whether fitting the class to a fully observed graph yields features useful for BO, and how this compares with using the true graph's smooth spectral modes at the same nominal dimension. The class arm fits the graph-valid surrogate using the entire graph, and the spectral arm extracts the $d$ smoothest true-graph Laplacian modes. We sweep $d\in\{4,8,16,32\}$ without changing the initialization size within a dataset.

BioPlex 293T uses the mean-essentiality objective on $13{,}167$ nodes, $N_0=20$, and $T=100$ subsequent BO rounds. DCSBM uses a nominal $5{,}000$ nodes before giant-component restriction, 16 generating communities, mean degree 20, and the latent objective, with $N_0=8$ and $T=100$, and the synthetic graph is generated separately for each seed. ESOL uses $1{,}117$ molecules, the real solubility objective, $N_0=32$, and $T=100$. ESOL is scored at $t=25$, before the full-information runs reach the regret floor. The configured UCB multiplier is $2$, with hyperparameter updates every 10 rounds. Full-graph access does not reveal unqueried objective values to BO.

\paragraph{Metric and aggregation.}
{Table~\ref{tab:exp-e6} reports the normalized simple regret of Section~\ref{sec:exp-setup} at the stated checkpoint, mean $\pm$ SEM over 20 seeds, with the complete objective used only for offline scoring.}

\begin{table}[t]
\centering\small
\caption{Full-information study: complete dimension sweep under full graph access. Entries are normalized simple regret, mean $\pm$ SEM over 20 seeds. The checkpoint $t$ counts BO rounds after initialization. The untruncated kernel is a separate reference tuned on a wider hyperparameter grid and is listed once per instance. At $d\le8$ on BioPlex both arms end on the same node in every seed, which gives identical cells.}
\label{tab:exp-e6}
\begin{tabular}{llcc}
\toprule
Instance / checkpoint & $d$ & Ours, full graph & Full-graph spectral\\
\midrule
BioPlex 293T & 4 & $.280\pm.000$ & $.280\pm.000$\\
($N_0=20$, $t=100$) & 8 & $.280\pm.000$ & $.014\pm.014$\\
 & 16 & $.112\pm.031$ & $.000\pm.000$\\
 & 32 & $.031\pm.022$ & $.052\pm.028$\\
\multicolumn{2}{l}{Untruncated reference} & \multicolumn{2}{c}{$.272\pm.039$}\\
\midrule
DCSBM, latent & 4 & $.231\pm.038$ & $.239\pm.034$\\
($N_0=8$, $t=100$) & 8 & $.194\pm.031$ & $.214\pm.032$\\
 & 16 & $.113\pm.023$ & $.077\pm.020$\\
 & 32 & $.118\pm.015$ & $.105\pm.025$\\
\multicolumn{2}{l}{Untruncated reference} & \multicolumn{2}{c}{$.147\pm.026$}\\
\midrule
ESOL & 4 & $.176\pm.016$ & $.176\pm.017$\\
($N_0=32$, $t=25$) & 8 & $.158\pm.031$ & $.118\pm.023$\\
 & 16 & $.139\pm.027$ & $.133\pm.021$\\
 & 32 & $.100\pm.025$ & $.062\pm.017$\\
\multicolumn{2}{l}{Untruncated reference} & \multicolumn{2}{c}{$.135\pm.024$}\\
\bottomrule
\end{tabular}
\end{table}

\paragraph{Reading.}
The sweep shows that class representations at dimensions far below the graph size are useful without the adjacency being low rank. On BioPlex the eigenbasis reaches its floor at $d=8$ while the class needs $d=32$, on DCSBM both arms improve up to $d=16$ and not beyond, and on ESOL the class improves across the sweep and stays above the eigenbasis. Full graph access removes the estimation of entries and leaves the class approximation, the solver, the kernel fit, and the policy in the comparison, so a difference between the two arms at fixed $d$ is a difference of representations. The dimension sweep is an offline sensitivity analysis: the partial-information algorithm never inspects the full graph to select $d$.

\subsection{Observation Patterns at Matched Entry Counts}
\label{app:exp-patterns}
The static observation-pattern study holds the estimator to our class surrogate and varies how a fixed number of unordered entries is observed: whole rows, uniformly sampled pairs, or a block followed by pinning. The feature map is then frozen for held-out GP prediction. Figure~\ref{fig:exp-patterns} reports rank correlation on ESOL and two dense in-class objectives at $d=16$, with 10 outer seeds. At 12,221 observed ESOL entries, the correlations are $0.373\pm0.014$ for rows, $0.138\pm0.007$ for uniform pairs, and $0.188\pm0.010$ for block-plus-pinning. At 124,785 entries they are $0.547\pm0.010$, $0.392\pm0.014$, and $0.517\pm0.014$, respectively. Thus the pattern of observations matters even when the number of entries is matched. Rows can deliver useful geometry with fewer entries in these settings, which helps explain why the two query types need separate acquisition studies. Among the pair patterns, a landmark block followed by probes against the landmarks, the pattern that Algorithm~\ref{Alg:pair} produces, is above uniform pairs at every entry count on all three instances, which is the static counterpart of the pair study of Section~\ref{sec:exp-q3}. The study measures held-out prediction from a frozen feature map, and Section~\ref{sec:exp-q3} gives the BO trajectories.

\begin{figure}[t]
\centering
\includegraphics[width=\linewidth]{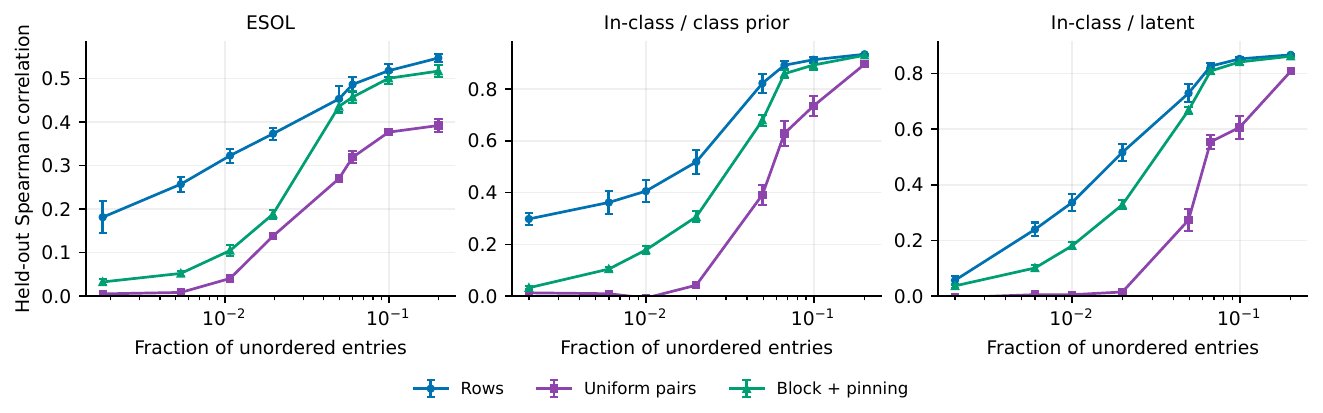}
\caption{Observation-pattern sensitivity with the class estimator, $d=16$. Held-out prediction Spearman correlation versus the fraction of distinct unordered entries, higher is better. Means and $\pm1$ SEM over 10 outer seeds. Each instance uses matched entry counts across the three displayed patterns.}
\label{fig:exp-patterns}
\end{figure}

\subsection{Pair Acquisition}
\label{app:exp-pairs}
\paragraph{Protocol.}
The ESOL pair study uses $n=1{,}117$, $d=N_0=32$, $T=200$, and 20 shared seeds. The 496 initial pairs are common to all three acquisition methods, which then purchase exactly $Q_E$ previously unobserved pairs before each objective evaluation. Ours (KG) combines landmark queries chosen by the KG rule with conditional membership updates and a pinning step capped at half the per-round budget. Ours (uniform pairs) samples uniformly from unrevealed unordered pairs and updates the class completion from those observations. Both use the structured-plus-bulk resolvent kernel, and this comparison evaluates their complete acquisition-and-update procedures rather than isolating the scoring rule. Local graph BO uses the same pair oracle and budget and builds its local context from the revealed entries. Its local-context threshold and restarts are part of the pair-oracle adaptation.

\paragraph{Budget response and optimization trajectories.}
Table~\ref{tab:exp-pair-round} reports all five budgets. Ours (KG) has the lowest mean final regret among the three acquisition methods throughout the sweep. Increasing $Q_E$ from 2 to 16 reduces its mean from $0.055$ to $0.011$, and the mean at 32 is $0.015$. Figure~\ref{fig:exp-pair-round} shows every trajectory, including the two reference methods. Random node search on the seeds of this study ends at $0.062\pm0.015$, and the full-graph spectral reference reaches zero final regret. Figure~\ref{fig:exp-acquisition}b shows the highest tested allocation.

\begin{table}[t]
\centering\small
\setlength{\tabcolsep}{7pt}
\caption{ESOL pair-query study: final normalized simple regret, mean $\pm$ SEM over 20 seeds, $d=N_0=32$, $T=200$. Each method acquires $Q_E$ pairs before every post-initialization evaluation. Bold marks the lowest mean among the three methods.}
\label{tab:exp-pair-round}
\begin{tabular}{rccc}
\toprule
$Q_E$ & Ours (KG) & Ours (uniform pairs) & Local graph BO\\\midrule
2 & $\mathbf{.055}\pm.012$ & $.072\pm.010$ & $.065\pm.014$\\
4 & $\mathbf{.047}\pm.011$ & $.055\pm.013$ & $.105\pm.011$\\
8 & $\mathbf{.040}\pm.011$ & $.059\pm.011$ & $.094\pm.010$\\
16 & $\mathbf{.011}\pm.007$ & $.037\pm.011$ & $.083\pm.016$\\
32 & $\mathbf{.015}\pm.008$ & $.037\pm.011$ & $.095\pm.012$\\
\bottomrule
\end{tabular}
\end{table}

\begin{figure}[t]
\centering
\includegraphics[width=\linewidth]{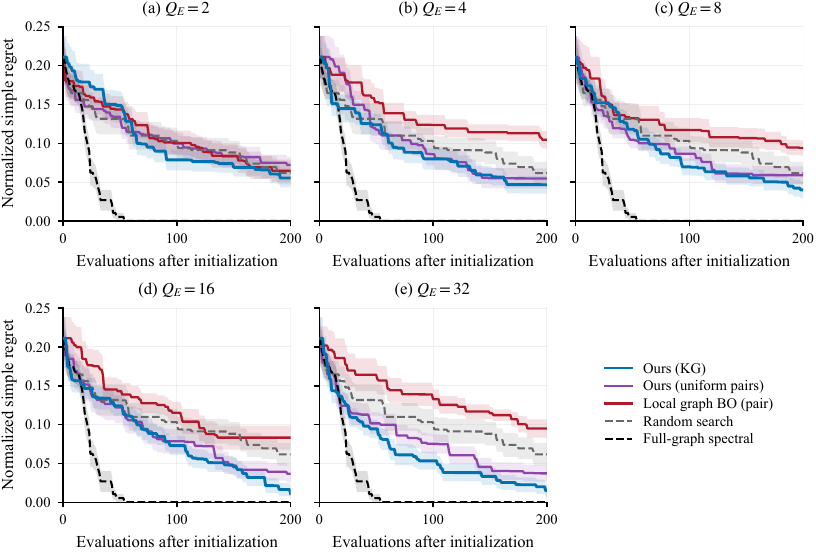}
\caption{ESOL pair-query study: all five acquisition budgets, mean $\pm1$ SEM over 20 seeds, $d=N_0=32$, $T=200$. The three acquisition methods share the initial 496 pairs and acquire $Q_E$ additional pairs per round. Dashed curves are random node search and full-graph spectral BO references, repeated across panels for context.}
\label{fig:exp-pair-round}
\end{figure}

\paragraph{Paired evidence and scope.}
The paired differences between Ours (KG) and local graph BO are negative with intervals excluding zero at $Q_E=4,8,16,32$ after the five-budget adjustment (Table~\ref{tab:exp-pair-ci}). Against uniform pairs all five paired differences favor Ours (KG). The comparison evaluates the acquisition-and-conditioning procedure as a whole, with the landmark geometry estimated from the available entries.

\begin{table}[t]
\centering\small
\setlength{\tabcolsep}{4pt}
\caption{Paired final-regret differences, Ours (KG) minus each comparator. Brackets are 99\% percentile bootstrap intervals, providing a five-budget Bonferroni adjustment within each comparator family. Negative differences favor Ours (KG). All pairs use the same 20 seeds.}
\label{tab:exp-pair-ci}
\begin{tabular}{rcc}
\toprule
$Q_E$ & Versus uniform pairs & Versus local graph BO\\\midrule
2 & $-.0169\;[-.0537,.0227]$ & $-.0093\;[-.0507,.0317]$\\
4 & $-.0081\;[-.0503,.0361]$ & $-.0581\;[-.0918,-.0244]$\\
8 & $-.0189\;[-.0510,.0130]$ & $-.0537\;[-.0851,-.0174]$\\
16 & $-.0255\;[-.0601,.0118]$ & $-.0720\;[-.1172,-.0250]$\\
32 & $-.0223\;[-.0495,.0020]$ & $-.0802\;[-.1130,-.0435]$\\
\bottomrule
\end{tabular}
\end{table}

\subsection{Row Rules}
\label{app:exp-rows}
The row-query study compares uniform rows with pinning, UCB-leader rows, and local-BFS rows using the same class learner on ESOL ($d=16$, $N_0=32$, $T=200$, 20 seeds). UCB-leader rows reveal the row of the unrevealed node with the largest score, and local BFS rows reveal rows in breadth-first order from the evaluated nodes along revealed edges. Table~\ref{tab:exp-row-rules} gives both $t=100$ and $200$, and Figure~\ref{fig:exp-row-round} gives the trajectories. {Neither alternative improves on the default consistently across budgets, and the differences are comparable to the SEM. Random node search on the seeds of this study gives $0.103\pm0.012$ at $t=100$ and $0.055\pm0.011$ at $t=200$, and every row rule improves on it at both checkpoints. At $t=200$ most configurations have many zero-regret runs, which limits discrimination between policies.}

\paragraph{Reading the row comparison.}
Rows provide common membership measurements across unrevealed nodes (Proposition~\ref{prop:shared}). With $N_0=32>d=16$, the local covariance is already zero whenever the fitted initial directions span the membership space. Further rows can still improve the surrogate fit on ESOL, where exact in-class identification is not assumed. The observed comparison supports pin-then-uniform as a practical default across these budgets.

\begin{table}[t]
\centering\small
\setlength{\tabcolsep}{4pt}
\caption{Row-query study: ESOL row policies, mean normalized regret $\pm$ SEM over 20 seeds, $d=16$, $N_0=32$. Additional rows at checkpoint $t$ are $\lfloor tQ_R\rfloor$.}
\label{tab:exp-row-rules}
\begin{tabular}{rrccc}
\toprule
$t$ & $Q_R$ & Uniform + pinning & UCB-leader & Local BFS\\\midrule
100 & 0.1 & $.047\pm.019$ & $.046\pm.017$ & $.050\pm.019$\\
100 & 0.25 & $.036\pm.015$ & $.028\pm.011$ & $.043\pm.016$\\
100 & 0.5 & $.016\pm.009$ & $.017\pm.012$ & $.031\pm.011$\\
100 & 1 & $.011\pm.008$ & $.027\pm.011$ & $.008\pm.008$\\
100 & 2 & $.000\pm.000$ & $.012\pm.008$ & $.000\pm.000$\\
\midrule
200 & 0.1 & $.006\pm.005$ & $.008\pm.006$ & $.011\pm.007$\\
200 & 0.25 & $.011\pm.011$ & $.005\pm.005$ & $.005\pm.005$\\
200 & 0.5 & $.005\pm.005$ & $.005\pm.005$ & $.000\pm.000$\\
200 & 1 & $.000\pm.000$ & $.000\pm.000$ & $.000\pm.000$\\
200 & 2 & $.000\pm.000$ & $.005\pm.005$ & $.000\pm.000$\\
\bottomrule
\end{tabular}
\end{table}

\begin{figure}[t]
\centering
\includegraphics[width=\linewidth]{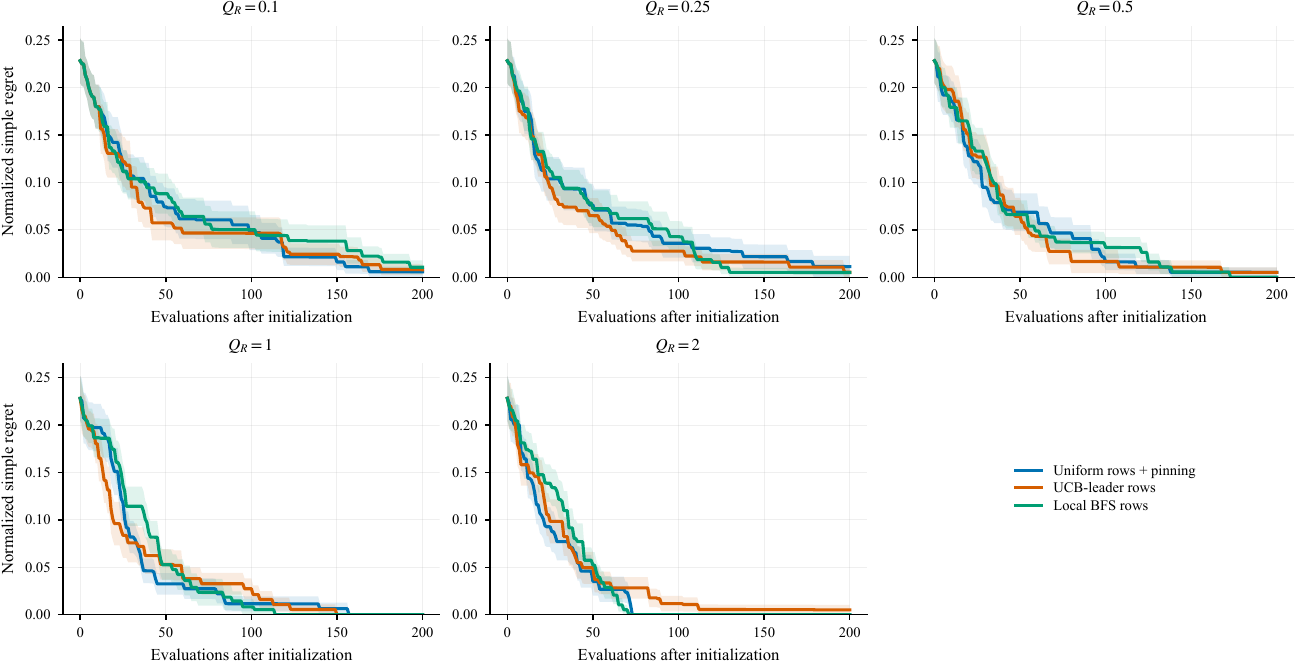}
\caption{Row-query study: ESOL row-rule trajectories, mean $\pm1$ SEM over 20 seeds, $d=16$, $N_0=32$, $T=200$. The additional row count at checkpoint $t$ is $\lfloor tQ_R\rfloor$. Later saturation motivates reporting both intermediate and final results. These are row-selection policies within Ours.}
\label{fig:exp-row-round}
\end{figure}

\section{Experiment Details}
\label{app:exp-details}

\subsection{Synthetic Random Graph Models}

\textbf{Dense in-class affinity.} The dense instance of Table~\ref{tab:results} is a member of the class. Each of $n=1{,}000$ nodes receives one of $d=16$ blocks uniformly at random and a degree factor $\theta_v\sim\mathrm{U}[0.5,1]$, its membership is $\pi_v=\theta_ve_{z_v}$, the core is $C=0.05\,\mathbf 1\mathbf 1^\top+0.25\,I$ (within-block affinity $0.30$, between-block $0.05$), and the affinity matrix is $\Pi C\Pi^\top$ with its diagonal removed, a weighted matrix with entries in $[0,1]$ that a row query reveals exactly. The regularizer $\zeta$ is the mean row sum.

\textbf{Sparse degree-corrected stochastic block model (DCSBM).} Each node $v$ receives a community label $z_v$ uniformly at random among $K=16$ communities and a degree factor $\theta_v=\exp(0.5\,g_v)$ with $g_v$ standard normal, truncated at $|g_v|\le2.5$ and normalized to mean one. The edge intensity between $u\ne v$ is $s\,\theta_u\theta_vB_{z_uz_v}$, where $B$ equals $10$ on the diagonal and $1$ elsewhere, and the scale $s$ is set so that the expected mean degree is $20$~\citep{karrer2011stochastic}. The graph is sampled block pair by block pair as a Poisson multigraph collapsed to a simple graph, so that no $n\times n$ array is formed, and it is restricted to its giant component. The intensity has a 16-community factorization, but collapsing Poisson counts to binary edges changes the expectation to $1-\exp(-s\theta_u\theta_vB_{z_uz_v})$ off the diagonal. The sampled binary graph is therefore an out-of-class instance for the rank-16 surrogate. We use $n\in\{2000,5000\}$.

\textbf{Sparse latent-position graph (RDPG).} Latent positions $x_v$ are uniform on $[0,1]^2$ and each pair is connected independently with probability $\rho\exp(-\|x_u-x_v\|^2/2h^2)$ with $h=0.25$, where $\rho$ is set so that the expected mean degree is $20$~\citep{young2007random}. The link is a fixed smooth kernel and not a shrinking geometric radius, so the expected adjacency has a smoothly decaying spectrum and the rank-$d$ class is an approximation with a nonzero bias. We use $n=5000$.

\subsubsection{Synthetic node functions}
\textbf{Latent.} On the DCSBM and on the dense in-class instance, $y(v)=a_{z_v}+0.5\,\epsilon_v$ with community levels $a_k$ and node noise $\epsilon_v$ standard normal. The objective has no degree term and is a function of the generating membership alone. On the RDPG, $y(v)=g(x_v)+4\exp(-\|x_v-c\|^2/2w^2)+0.5\,\epsilon_v$, where $g$ is a standardized random trigonometric polynomial with four terms of unit integer frequency, the bump has width $w=0.2$, and its centre $c$ is uniform on $[0.2,0.8]^2$. \textbf{Class-prior.} A draw from the GP prior of the structured-plus-bulk kernel built from the expected adjacency, plus independent noise whose variance is five percent of the total. All objectives are standardized to zero mean and unit variance. The parameters of the objectives were fixed on the design seeds.

\subsection{Real-World Graph Models}
\textbf{BioPlex and DepMap.} BioPlex 3.0~\citep{huttlin2021dual} reports protein interactions detected by affinity-purification mass spectrometry in HEK293T and HCT116 cells. We use each network as an undirected binary graph restricted to its largest connected component and to the genes present in the DepMap CRISPR gene-effect data~\citep{tsherniak2017defining}, which gives $n=13{,}167$ proteins with mean degree $16.4$ for 293T and $n=9{,}609$ for HCT116, of which $9{,}494$ have a gene effect measured in the HCT116 screen and form the node set of the own-line objective. The objective is the negated DepMap CRISPR gene effect of the gene encoding each protein, so that larger values mean stronger dependency, either averaged over all screened cell lines (mean essentiality) or taken in a single line: C2BBe1 (DepMap ACH-000009, colon adenocarcinoma), OPM-2 (ACH-000024, multiple myeloma), and, for the HCT116 network, HCT116 itself (ACH-000971). The two lines were fixed by their DepMap identifiers before any run, and the gene effects are those of DepMap release 24Q4 (the release name of the fourth quarter of 2024). A structural query corresponds to one bait experiment and a function evaluation to one knockout phenotype.

\textbf{ESOL.} The ESOL collection~\citep{delaney2004esol} provides measured aqueous solubilities, of which we use $1{,}117$ molecules. The affinity is the Tanimoto similarity of Morgan fingerprints of radius two, a dense matrix with entries in $[0,1]$. The Tanimoto similarity is cheap to compute and stands in for expensive pairwise similarities, for which each entry carries a real cost.

\textbf{Chameleon.} The chameleon page graph of the MUSAE collection~\citep{rozemberczki2021multi} has Wikipedia articles as nodes and mutual hyperlinks as edges. We symmetrize and binarize the graph, keep its largest connected component, collapse nodes with identical neighbor sets onto one representative, and keep the largest connected component again, which leaves $n=1{,}223$ nodes. The objective is the logarithm of the average monthly traffic of the page.

\textbf{Scope of the benchmark.} The tasks cover dense affinities, sparse biological networks, and synthetic models. Different BioPlex objectives share a network, so their results are reported separately and never pooled into one significance test.

\subsection{Baseline Algorithms and Implementation Settings}
\label{app:exp-baselines}
\textbf{Random node search} evaluates a uniformly sampled unevaluated node and buys no additional structure. {\textbf{Local search} evaluates a revealed, unevaluated neighbor of its current node, moves when the value improves, discovers neighborhoods through the same charged oracle, and restarts at a uniformly drawn unevaluated node when its accessible neighborhood is exhausted.} \textbf{Initial structure only} runs our procedure without buying additional rows after initialization and still updates the function posterior and kernel hyperparameters. \textbf{Ours, full graph} fits the surrogate with full structural access, whereas \textbf{full-graph spectral} uses the true graph's low-dimensional spectral features. \textbf{Untruncated} uses the full-spectrum kernel. These three references are distinct and are labeled separately throughout.

{\textbf{Local graph BO}~\citep{wan2023bayesian} is the \texttt{poly-default} configuration of the published method. In the row studies it runs on the original graph with neighborhood access charged against the same cumulative row allowance as Ours, and in the pair study it is adapted to the pair oracle, where a local step cannot obtain an unseen neighborhood for free and restarts are part of the adaptation. Apart from the disclosed $k$NN view on the dense and ESOL affinities it receives no uncharged full-graph preprocessing.} On those dense affinities every revealed row is complete, and local search uses the same $k$NN view to define neighborhoods.

{\textbf{Acquisition controls} retain the class learner and vary the structural queries: uniform pairs against Ours (KG) in the pair study, and uniform rows with pinning, UCB-leader rows, and local-BFS rows in the row-rule study. Membership conditioning differs between the two pair arms, so that comparison is a comparison of procedures.}

Our main row configurations use $d=16$, a structured-plus-bulk resolvent kernel, UCB multiplier $\kappa=2$, refits and hyperparameter updates every 10 evaluations, and pinning share $\rho_{\rm pin}=3/4$. The multiplier $\kappa=2$ is the usual practical choice and is smaller than the union-bound schedule of Assumption~\ref{ass:confidence}. The ESOL pair studies use $d=32$, $N_0=32$, and $\zeta=1$. Initialization, horizon, and structural schedule vary by study as specified in Table~\ref{tab:exp-protocols}. The full-information study changes $d$ at fixed initialization within each dataset, and the scale ablation below uses its separately held-out evaluation seeds. Configuration files and numerical-source manifests are preserved in the experiment archive.

\section{Ablation Study}
\label{sec:ablation}
{This appendix reports the kernel-scale ablation. Resolution, budgets, observation patterns, and acquisition rules are in Appendix~\ref{app:exp}.}

\subsection{Structured and Bulk Scales}
\label{sec:ablation-scales}
This ablation asks whether the exact one-scale resolvent kernel $r(\tilde L_\zeta)$ of Eq.~\eqref{Equ:kernel} is a valid alternative to the structured-plus-bulk kernel of Eq.~\eqref{Equ:kernel2}. It compares the two using the same surrogate family, under row queries with the row rule of Section~\ref{Subsec:edge design}, on BioPlex 293T with the mean-essentiality objective ($N_0=20$) and on the DCSBM with $n\approx5{,}000$ and the latent objective ($N_0=8$), both at resolution $16$ and at $250$ and $1{,}000$ total revealed rows. The ten evaluation seeds are disjoint from the design seeds on which the default kernel was selected, and the two kernels use paired cold starts and matched structural budgets. Figure~\ref{fig:scales} shows the regret trajectories and the contrast diagnostic, and Table~\ref{tab:scales} the final normalized regret after $100$ evaluations. The two kernels perform comparably: none of the four paired differences is significant at the $5\%$ level, the medians are split, and the means differ by amounts comparable to the SEM. The one-scale kernel is therefore a valid alternative, and the main text uses the structured-plus-bulk kernel because it supports the linear representation for every core and separates the structured and bulk scales.

The ratio $\operatorname{std}_v\mu_t(v)/\operatorname{mean}_v\kappa s_t(v)$, the variation of the structured posterior mean relative to the exploration term, is about one order of magnitude larger for the structured-plus-bulk kernel than for the one-scale kernel on both instances and both budgets (Figure~\ref{fig:scales}, bottom). Decoupling the two scales thus raises the structured contrast of the acquisition, as the dilution argument of Appendix~\ref{app:method} predicts, while the regret trajectories overlap.
 The one-scale resolvent retains all $16$ modes and the structured-plus-bulk kernel $9$ to $11$, so the comparison changes retained directions as well as the structured-to-bulk contrast.

\begin{table}[t]
\centering\small
\caption{One-scale against structured-plus-bulk resolvent kernels: final normalized regret after $100$ evaluations, mean $\pm$ SEM and median over $10$ paired evaluation seeds, wins, losses and ties of the one-scale kernel, and the sign-test $p$-value. Budgets are total revealed rows.}
\label{tab:scales}
\begin{tabular}{llrrrrrr}
\toprule
Instance & Rows & One-scale & Two-scale & Median one & Median two & W/L/T & $p$\\
\midrule
BioPlex 293T & 250 & $.300\pm.037$ & $.365\pm.054$ & $.312$ & $.364$ & 4/1/5 & $.375$\\
BioPlex 293T & 1,000 & $.365\pm.038$ & $.377\pm.054$ & $.312$ & $.436$ & 5/4/1 & $1.000$\\
DCSBM & 250 & $.151\pm.043$ & $.176\pm.045$ & $.140$ & $.127$ & 5/3/2 & $.727$\\
DCSBM & 1,000 & $.142\pm.024$ & $.169\pm.043$ & $.144$ & $.141$ & 5/2/3 & $.453$\\
\bottomrule
\end{tabular}
\end{table}

\begin{figure}[t]
\centering
\includegraphics[width=\linewidth]{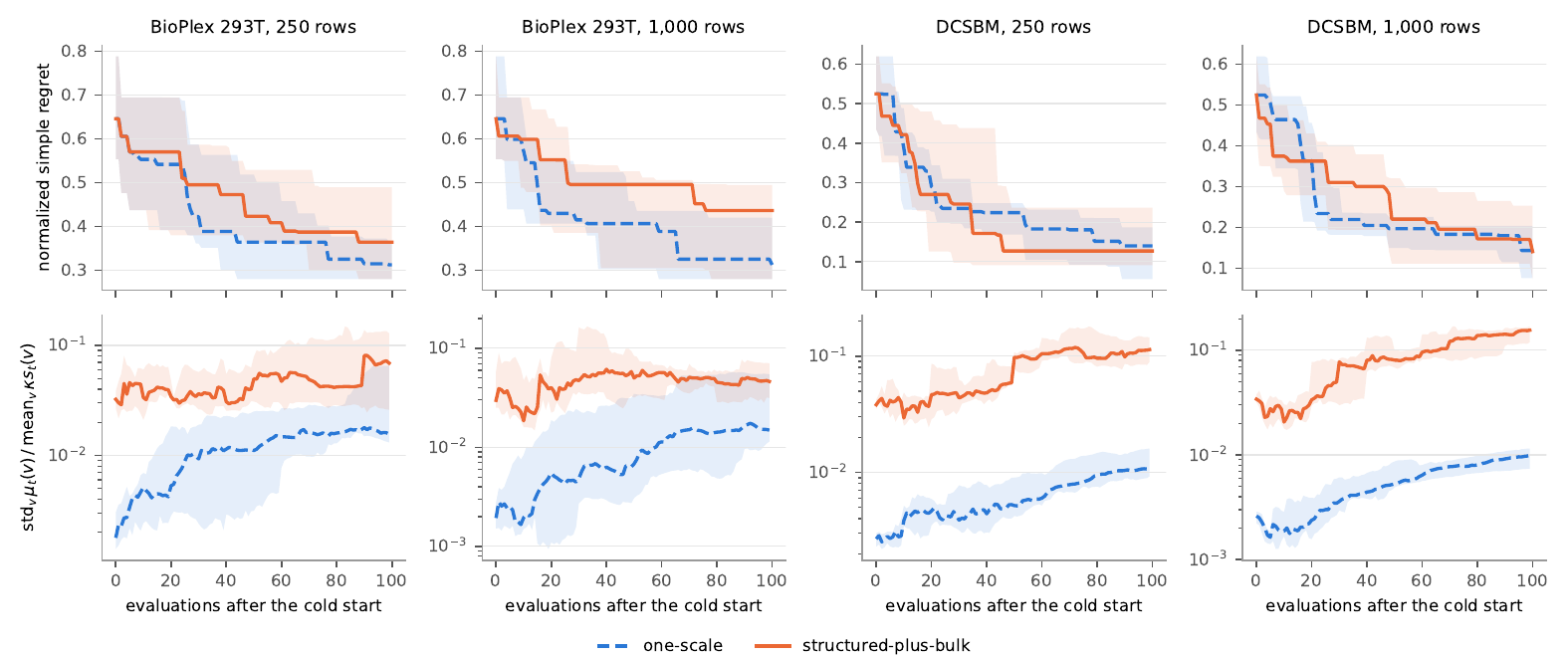}
\caption{One-scale against structured-plus-bulk resolvent kernels using the same surrogate family, row queries, resolution $16$, median with interquartile band over $10$ paired evaluation seeds. Top: normalized simple regret against evaluations after the cold start. Bottom: the structured contrast of the acquisition, $\operatorname{std}_v\mu_t(v)/\operatorname{mean}_v\kappa s_t(v)$, on a logarithmic axis. For the resolvent shape, decoupling the scales raises the contrast by about one order of magnitude on these instances while the regret trajectories overlap.}
\label{fig:scales}
\end{figure}

\endgroup
\end{document}